\documentclass[10pt,journal,compsoc]{IEEEtran}
\usepackage{amsmath,amsfonts}
\usepackage{algorithmic}
\usepackage{algorithm}
\usepackage{array}
\usepackage{textcomp}
\usepackage{stfloats}
\usepackage{url}
\usepackage{verbatim}
\usepackage{graphicx}
\usepackage{cite}
\usepackage{xspace}
\usepackage{threeparttable}
\usepackage{xcolor}
\usepackage{multirow}
\usepackage[T1]{fontenc}
\usepackage{gensymb}
\usepackage{booktabs}
\usepackage{amssymb}
\usepackage{enumitem}
\usepackage[hyphenbreaks]{breakurl}
\usepackage{mathtools}
\usepackage{xspace}
\usepackage{diagbox}
\usepackage{graphbox}
\usepackage{colortbl}
\usepackage{graphicx}
\usepackage{subcaption}
\usepackage{pifont}
\usepackage{makecell}

\usepackage[table]{xcolor} 
\usepackage{colortbl}      
\usepackage[colorlinks=true, linkcolor=blue, citecolor=red, urlcolor=magenta]{hyperref}
\definecolor{mygray}{gray}{0.9} 
\definecolor{myblue}{RGB}{214,224,240} 
\definecolor{mygreen}{RGB}{214,240,225} 
\definecolor{mywhite}{RGB}{255,255,255} 

\definecolor{PerceptionBG}{RGB}{216,232,248} 
\definecolor{CognitionBG}{RGB}{249,228,196}  
\definecolor{ActionBG}{RGB}{219,236,210}     

\usepackage{pifont}
\newcommand{\cmark}{\textcolor{green!60!black}{\ding{51}}}
\newcommand{\xmark}{\textcolor{red}{\ding{55}}}

\usepackage[capitalize]{cleveref}
\crefname{section}{Sec.}{Secs.}
\Crefname{section}{Section}{Sections}
\Crefname{table}{Table}{Tables}
\crefname{table}{Tab.}{Tabs.}

\hypersetup{hidelinks} 

\makeatletter
\newcommand*{\rom}[1]{\expandafter@slowromancap\romannumeral #1@}
\makeatother

\graphicspath{ {./figure/} }

\hypersetup{
    colorlinks=true,
    linkcolor=black,
    citecolor=green,
    urlcolor=magenta
}

\ifCLASSINFOpdf
\else
\fi

\begin{document}

\title{Advancing Adaptive Multi-Stage Video Anomaly Reasoning: A Benchmark Dataset and Method}

\author{Chao Huang, Benfeng Wang, Wei Wang, Jie Wen, Li Shen, Wenqi Ren, Yong Xu, Xiaochun Cao
        
\IEEEcompsocitemizethanks{\IEEEcompsocthanksitem C. Huang, B. Wang, W. Wang, L. Shen, W. Ren, and X. Cao are with the School of Cyber Science and Technology, Shenzhen Campus of Sun Yat-sen University, Shenzhen 518107, China (e-mail: \{huangch253, shenli6, renwq, caoxiaochun\}@mail.sysu.edu.cn, wangbf23@mail2.sysu.edu.cn).
\IEEEcompsocthanksitem J. Wen and Y. Xu are with the Shenzhen Key Laboratory of Visual Object Detection and Recognition, Harbin Institute of Technology, Shenzhen, 518055, China (e-mails: Jiewen\_pr@126.com, laterfall@hit.edu.cn).
\protect



}
}

\markboth{Submitted to IEEE Transactions on Pattern Analysis and Machine Intelligence}%
{Shell \MakeLowercase{\textit{et al.}}: Bare Demo of IEEEtran.cls for Computer Society Journals}

\IEEEtitleabstractindextext{%
\begin{abstract}

Recent progress in reasoning capabilities of Multimodal Large Language Models(MLLMs) has highlighted their potential for performing complex video understanding tasks. However, in the domain of Video Anomaly Detection and Understanding (VAD\&U), existing MLLM-based methods are largely limited to anomaly localization or post-hoc description, lacking explicit reasoning processes, risk awareness, and decision-oriented interpretation. To address this gap, we define a new task termed Video Anomaly Reasoning (VAR), which elevates video anomaly analysis from descriptive understanding to structured, multi-stage reasoning. VAR explicitly requires models to perform progressive reasoning over anomalous events before answering anomaly-related questions, encompassing visual perception, causal interpretation, and risk-aware decision making. To support this task, we present a new dataset with 8,641 videos, where each video is annotated with diverse question types corresponding to different reasoning depths, totaling more than 50,000 samples, making it one of the largest datasets for video anomaly. The annotations are based on a  structured \textbf{Per}ception–\textbf{Co}gnition–\textbf{Act}ion \textbf{C}hain-of-\textbf{T}hought (PerCoAct-CoT), which formalizes domain-specific reasoning priors for video anomaly understanding. This design enables systematic evaluation of multi-stage and adaptive anomaly reasoning. In addition, we propose Anomaly-Aware Group Relative Policy Optimization (A\textsuperscript{2}-GRPO) to further enhance reasoning reliability under weak supervision. Building upon the proposed task and dataset, we develop an end-to-end MLLM-based VAR model termed Vad-R1-Plus, which supports adaptive hierarchical reasoning and risk-aware decision making. Extensive experiments demonstrate that the proposed benchmark and method effectively advance the reasoning capabilities of MLLMs on VAR tasks, outperforming both open-source and proprietary baselines. Our datasets and models will be made publicly available at \url{https://github.com/wbfwonderful/Vad-R1-Plus}.

\end{abstract}

}

\maketitle

\IEEEdisplaynontitleabstractindextext

%
\IEEEpeerreviewmaketitle

\IEEEraisesectionheading{\section{Introduction}\label{plus-sec:introduction}}

\IEEEPARstart{T}{he} emergence of reasoning capabilities in Large Language Models (LLMs) represents a significant milestone in the development of foundation models, where models are required to explicitly think before answering. The remarkable performance of OpenAI o1~\cite{jaech2024openai} highlights that explicit reasoning can greatly enhance causal analysis and deductive capabilities. Importantly, the open-source DeepSeek-R1~\cite{guo2025deepseek} provides evidence that carefully designed reinforcement-based optimization strategies can reliably elicit stronger reasoning abilities. Building on these insights, recent research has begun extending reasoning-enhancement techniques from the text-only domain to the multimodal domain, enabling Multimodal Large Language Models (MLLMs) to perform better on a wide range of visual tasks such as grounding, mathematical reasoning, and OCR~\cite{team2025kimi, QwenQvQ}.

\begin{figure}[t]
\centering
\includegraphics[width=1\columnwidth]{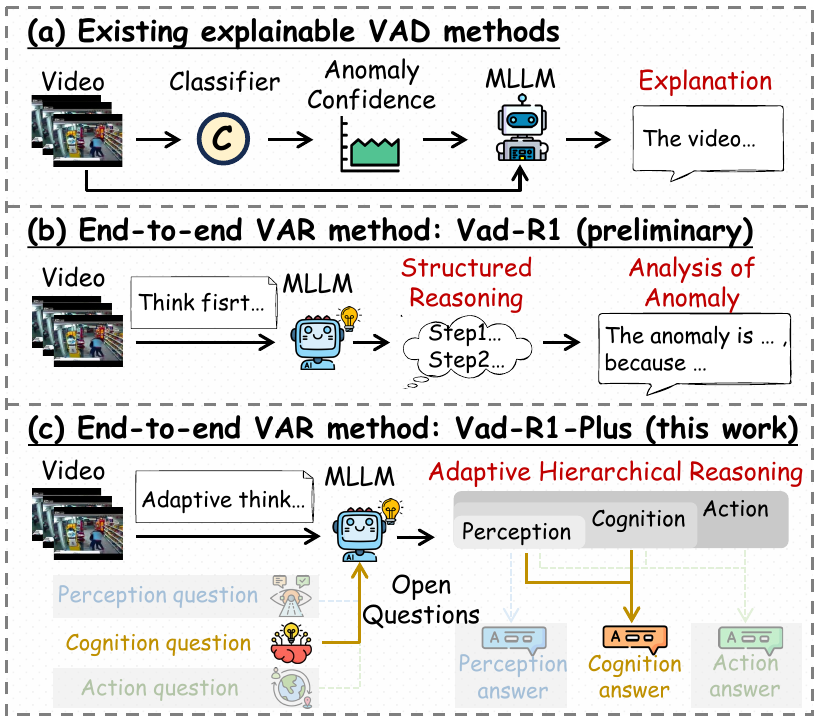} 
\caption{Task comparison. (a) Existing explainable VAD methods typically output an additional explanation after anomaly confidence. (b) Our preliminary method Vad-R1 could conduct deep reasoning about abnormal events. (c) This paper presents Vad-R1-Plus, which supports adaptive hierarchical reasoning and answer open questions.}
\label{plus-fig:motivation}
\end{figure}

Among these tasks, Video Anomaly Detection and Understanding (VAD\&U) is particularly challenging. Abnormal events in real-world videos often involve dynamic human behaviors and complex, unconstrained environments. Analyzing abnormal events is not merely a matter of identifying unusual patterns, but also requires a deeper understanding of what has happened, why it is abnormal, and how it may influence public safety and decision-making.

Traditional VAD methods usually fall into two paradigms: semi-supervised~\cite{luo2021future, huang2024long, huang2022self, cao2024scene, huang2021abnormal} and weakly-supervised VADs~\cite{wu2020not, sultani2018real, huang2022weakly, huang2021abnormal,lv2021localizing, huang2025multimodal}. These methods typically predict whether a video contains abnormal frames or segments, lacking understanding and explanation of anomalies. To enhance explainability, some studies focus on explainable VAD methods~\cite{lv2024video, zhang2024holmes, zhang2024holmesvau, huangexvad, cai2025hiprobe}. As shown in Figure~\ref{plus-fig:motivation}(a), these pipelines first estimate frame or segment-level anomaly confidence with a classifier, then an MLLM generates supplementary explanations. In such pipelines, reasoning is detached from detection, and the generated explanations do not influence and enrich the model’s understanding of the abnormal event. More recent works employ MLLMs directly for anomaly question answering and description~\cite{tang2024hawk, ma2025sherlock}, lacking thinking and analytical abilities. More importantly, these methods lack the ability of risk awareness and decision-oriented interpretation, which are crucial in realistic applications. A model that only identifies an abnormal event but fails to assess its severity and suggest appropriate actions remains insufficient for real-world deployment.

To bridge this gap, we introduce the new task of Video Anomaly Reasoning (VAR), which moves beyond the traditional detection paradigm. As shown in Figure~\ref{plus-fig:motivation}(b) and (c), compared with existing explainable VAD methods where anomaly detection and explanation are two separate steps, our VAR is an end-to-end task. This design requires MLLMs to perform structured, multi-stage reasoning before answering, aligning with the human cognitive process, enabling contextual understanding, behavior interpretation, and norm violation analysis. In this unified pipeline, anomaly detection and explanation are integrated in a single forward pass and can mutually reinforce each other. 

Realizing video anomaly reasoning in realistic scenarios requires both appropriate supervision and principled evaluation. Existing video anomaly datasets are primarily designed for detection or coarse-grained understanding, lacking explicit reasoning annotations and diverse question formulations. As a result, they are insufficient for training and evaluating models on structured, multi-stage anomaly reasoning behaviors. To address this limitation, we construct \textbf{Vad-Reasoning-Plus}, a large-scale benchmark dataset specifically tailored for the VAR task with 8,461 videos. To better reflect real-world usage, where users may raise heterogeneous questions targeting different aspects of an abnormal event, the dataset includes three categories of questions corresponding to different reasoning stages. Each category encourages a distinct depth of reasoning, and multiple question formulations are designed within each category to cover diverse sub-tasks and perspectives. This multi-level question design exposes models to mixed and adaptive reasoning demands that naturally arise in real-world anomaly understanding scenarios. In total, Vad-Reasoning-Plus contains more than 50,000 samples, making it one of the largest datasets for video anomaly tasks.

The construction of Vad-Reasoning-Plus is guided by a structured multi-stage \textbf{Per}ception–\textbf{Co}gnition–\textbf{Act}ion Chain-of-Thought (PerCoAct-CoT), which formalizes domain-specific reasoning priors underlying video anomaly understanding. Aligned with the human cognitive process of interpreting abnormal events, PerCoAct-CoT decomposes anomaly reasoning into three explicit stages. In the \textbf{Perception} stage, models are guided to observe the video context and identify suspicious visual cues. In the \textbf{Cognition} stage, models perform anomaly interpretation, causal analysis, and category judgment. Finally, in the \textbf{Action} stage, models assess potential risks and generate decision-oriented responses. By organizing annotations around these stages, the dataset promotes structured and interpretable reasoning instead of unconstrained free-form explanations.

For the open-ended VAR task, it is difficult to define clear training objectives. To address this, we adopt a progressive training paradigm with two stages. We firstly perform Supervised Fine-Tuning (SFT) to equip MLLM with basic anomaly reasoning capabilities. Then Reinforcement Learning (RL) is conducted with the proposed Anomaly Aware Group Relative Policy Optimization (A\textsuperscript{2}GRPO) algorithm, an extension of GRPO~\cite{shao2024deepseekmath} with anomaly verification, risk assessment and reasoning depth punishment rewards, which enhance MLLM's video anomaly reasoning capabilities with only video-level weak annotations. Finally, we build Vad-R1-Plus, a dedicated VAR model equipped with adaptive hierarchical reasoning and risk-aware decision-making capabilities. In summary, the contributions of this paper are summarized as follows:

\begin{itemize}
    \item We define the new task of Video Anomaly Reasoning (VAR) , which moves beyond traditional VAD\&U, requiring the models to perform structured, multi-stage reasoning before predicting the results.
    \item We construct Vad-Reasoning-Plus, a large-scale dataset tailored for video anomaly reasoning, which contains explicit reasoning annotations and three categories of questions aligned with different reasoning stages.
    \item We propose a structured Perception-Cognition-Action Chain-of-Thought, a domain-specific multi-stage reasoning paradigm, which decomposes anomaly reasoning into Perception, Cognition, and Action stages, explicitly introducing the domain priors underlying video anomaly tasks to models.
    \item We propose Anomaly Aware GRPO, a RL algorithm tailored for VAR, which extends GRPO with three special designed rewards, enhancing anomaly reasoning capabilities with limited annotations.
    \item Extensive experiments demonstrate that our Vad-R1-Plus model achieves state-of-the-art performance, significantly outperforming both open-source and proprietary MLLMs under multiple evaluation metrics.
\end{itemize}

This paper is an extension of our previous work~\cite{huang2025vadr1}. The enhancements of this paper are as follows:

\begin{itemize}
    \item We extend the original dataset to a larger scale. The new Vad-Reasoning-Plus dataset contains three categories of open-ended questions with different reasoning depth, aligning with the proposed PerCoAct-CoT.
    \item We augment the original chain-of-thought by introducing a new Action stage, enabling the model to assess risks and provide decision-oriented suggestions.
    \item We introduce A\textsuperscript{2}GRPO with new risk assessment and reasoning depth punishment rewards to improve risk-aware and adaptive reasoning capabilities.
    \item We present a stronger model, Vad-R1-Plus, which is equipped with adaptive hierarchical reasoning and risk-aware decision-making capabilities.
    \item We conduct more comprehensive experiments, including broader baselines, deeper ablations, and more detailed assessments of reasoning quality.
\end{itemize}

\section{Related Works}
\subsection{Video Anomaly Detection}


Traditional VAD has been extensively explored under semi-supervised and weakly supervised paradigms. Early methods mainly focused on the semi-supervised paradigm, which distinguishes anomalies from normal patterns by learning regularities of visual features~\cite{ luo2021future, huang2024long, huang2022self, huang2021abnormal, cao2024scene}. While semi-supervised VAD methods rely on modeling normal patterns, they often struggle with the limited coverage of normal training data and the diversity of real-world anomalies. Consequently, research has shifted towards weakly supervised VAD, where models are trained with only coarse video-level labels to reduce annotation costs~\cite{wu2020not, sultani2018real, huang2022weakly, huang2021abnormal, lv2021localizing, huang2025multimodal}. Based on the pre-trained vision-language models, some studies try to introduce textual description to enhance detection~\cite{wang2025federated, wu2024vadclip, chen2023tevad}. Furthermore, some studies use carefully designed pipelines and collaborative multimodel frameworks to achieve training-free methods~\cite{zanella2024harnessing, yang2024follow, li2025vadtree, shao2025eventvad}. To enhance interpretability, explainable VAD methods combine detection with post-hoc descriptions generated by MLLMs~\cite{ye2025vera, huangexvad, lv2024video}. More recently, several works attempt to use MLLMs in an end-to-end manner, directly generating answers or descriptions for anomaly-related queries. HAWK~\cite{tang2024hawk} introduces motion information into MLLM to promote anomaly understanding and question answering. Sherlock~\cite{ma2025sherlock} achieves efficient video anomaly extraction and localization through structured video semantic information. Holmes-VAU~\cite{zhang2024holmesvau} proposes an anomaly-focused temporal sampler to guide MLLM for better understanding of abnormal events. A-Guardian~\cite{du2024uncovering} enhances causal understanding of video anomalies through a dual-prompt mechanism. VA-GPT~\cite{chen2025aligning} selects important informative spatial tokens and generates anomaly-aware temporal tokens to enable MLLM to focus on anomaly regions and moments. However, these studies remain at shallow understanding with MLLMs, lacking in-depth exploration of reasoning capability. A small but emerging line of research explores reasoning for video anomalies, incorporating step-by-step thinking, causal explanation, or temporal logic~\cite{huang2025vadr1, zhu2025vau, yu2025cuebench, mo2025a2seek}. However, these methods lack risk-aware analysis and decision-oriented recommendations, which are essential for moving toward agent-like anomaly understanding.

\subsection{Video Anomaly Dataset}

The existing VAD datasets primarily provide coarse-grained category labels~\cite{sultani2018real, wu2020not, lv2021localizing, acsintoae2022ubnormal}, which are sufficient for anomaly detection but inadequate for understanding the nature and causes of abnormal events. To address this limitation, CUVA~\cite{du2024uncovering} and ECVA~\cite{du2024exploring} provide descriptions, reasons and effects of anomaly. HIVAU-70k~\cite{zhang2024holmesvau} provides multi-granularity descriptions to enable complex anomaly analysis. Besides, other datasets~\cite{yuan2024towards, tang2024hawk, zhao2025smarthome, liu2025surveillancevqa, kim2025vru} also provide descriptions and question-answer pairs for abnormal events. These efforts make important steps towards enhancing the understanding of the abnormal event in the videos. However, they still lack structured reasoning annotations.


\subsection{Video Multimodal Large Language Model}

The video multimodal large models provide an interactive way to understand video content. Early works like VideoChat~\cite{li2023videochat} integrate visual encoders into large language models by aligning visual and textual tokens via mapping networks, while Video-LLaVA~\cite{lin2023video} jointly train the projector layers with both images and videos. Compared to static images, videos contain more redundant and temporally correlated information. Consequently, some studies explore token compression mechanisms to obtain longer context. LLaMA-VID~\cite{li2024llama} represents each video frame with two tokens through computing context attention between text and video frames. Chat-univi~\cite{jin2024chat} proposes a dynamic token merging strategy and applies clustering algorithms at the token level to reduce duplicate input tokens, while LongVA~\cite{zhang2024long} proposes a strategy for long context transfer, which first extends the context on the text model and then directly transfers it to the visual domain without long video training. In addition, recent works have explored online video stream understanding. StreamChat~\cite{xiong2025streaming} utilize a hierarchical memory system to effectively process and compress video features, achieving real-time and multi-turn dialogue. ReKV~\cite{di2025streaming} enables real-time and accurate responses through sliding-window attention and in-context KV-Cache retrieval to efficiently preserve long-term context. Nevertheless, these methods remain primarily at the level of video understanding and lack systematic exploration of explicit reasoning capabilities.

\subsection{Reasoning Capability for MLLMs}

Recently, enhancing the reasoning capability of MLLMs has become a major research focus. Some studies propose multi-stage reasoning frameworks and large-scale CoT datasets to enhance the reasoning capability of MLLMs. LLaVA-o1~\cite{xu2024llava} designs a reasoning process with four stages and employ a stage-level beam search strategy to improve the reasoning capability of MLLMs. LlamaV-o1~\cite{thawakar2025llamav} optimizes reasoning paths, enabling effective reasoning capability in complex multi-step visual reasoning tasks. VideoMind~\cite{liu2025videomind} proposes four collaborative agents with a chain-of-LoRA strategy, making it seamless for role switching. Furthermore, some studies adopt training-free strategies such as latent intervention or visual-context-injected CoT~\cite{zhan2025l2v}, which introduce reasoning patterns without retraining. Additionally, inference-time search techniques like stage-level beam search or tree-of-thought reasoning have shown to boost consistency and robustness. Recently, DeepSeek-R1~\cite{guo2025deepseek} demonstrates the potential of reinforcement learning in enhancing the reasoning capability, inspiring subsequent efforts to reproduce its success in multimodal domains. Vision-R1~\cite{huang2025vision} first explores how RL can improve the reasoning capability of MLLMs. In the field of video, some studies also utilize RL to improve spatial reasoning~\cite{li2025videochat}, temporal reasoning~\cite{wang2025timezero} and general causal reasoning~\cite{feng2025video}.

\section{Method}

This section describes the key components of our method for video anomaly reasoning. Specifically, we first introduce the Vad-Reasoning-Plus dataset in Section~\ref{plus-sec:dataset}, which is designed to support adaptive hybrid reasoning across different anomaly-related queries. The dataset is organized according to a structured Perception–Cognition–Action Chain-of-Thought (PerCoAct-CoT) as described in Section~\ref{plus-sec:cot}, which formalizes domain-specific reasoning priors in video anomaly understanding. Finally, we present an anomaly-aware reinforcement learning algorithm, A\textsuperscript{2}-GRPO in Section~\ref{plus-sec:grpo}, which is used to further enhance reasoning reliability under weak supervision.

\begin{figure*}[t]
    \centering
    \includegraphics[width=1\linewidth]{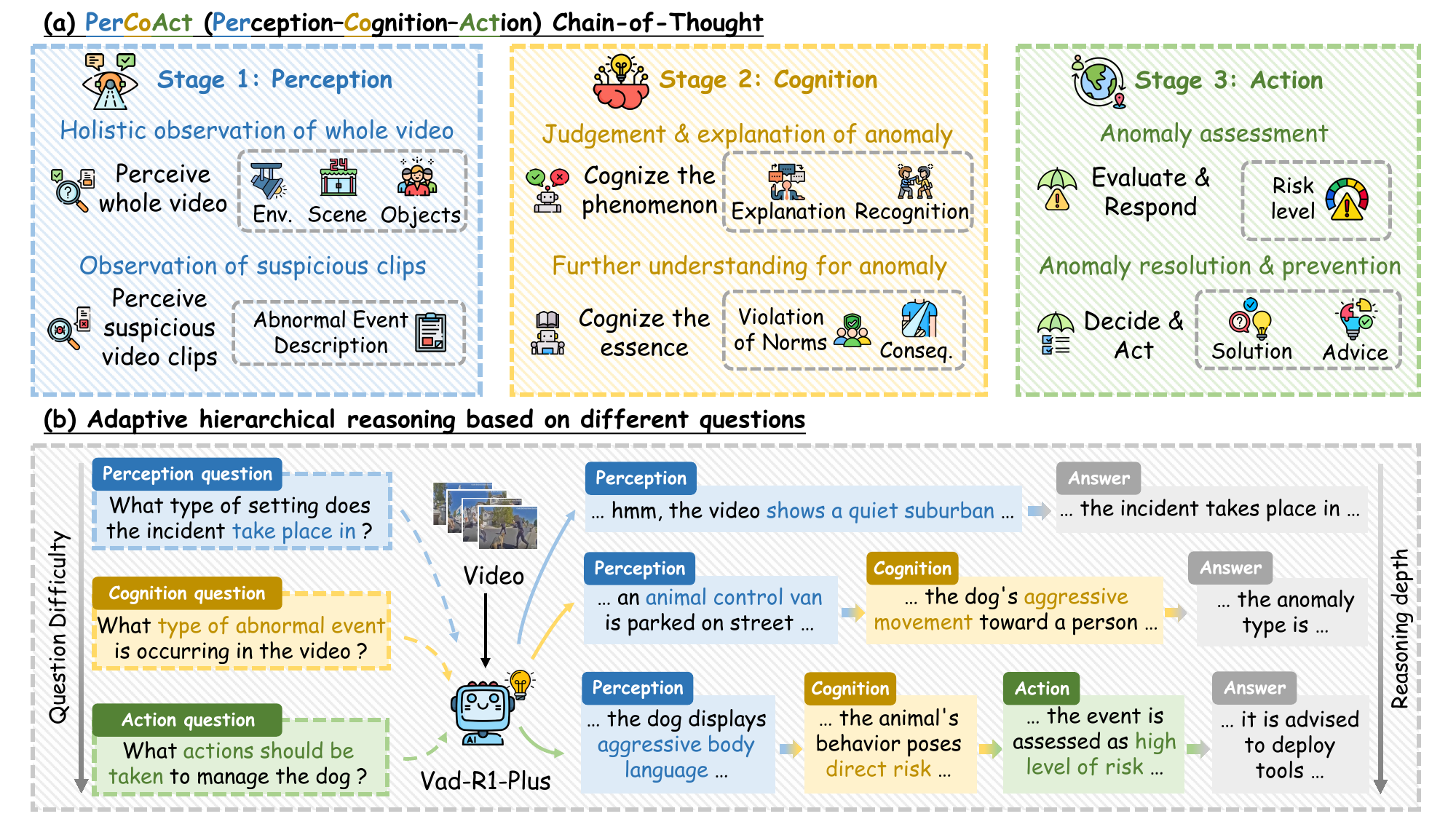} 
    \caption{Illustration of the proposed structured Chain-of-Thought and the adaptive hybrid reasoning mechanism.}
    \label{plus-fig:cot}
\end{figure*}

\subsection{Vad-Reasoning-Plus Dataset}\label{plus-sec:dataset}

\subsubsection{Adaptive Hybrid Reasoning}

The design of the Vad-Reasoning-Plus dataset is motivated by the observation that real-world anomaly-related queries naturally require different depths of reasoning. In real-world applications, user queries are often highly diverse, where some questions require only basic perceptual descriptions of the video, while others demand deeper causal analyses or even explicit safety-oriented decisions. Treating all questions uniformly with a fixed-depth reasoning process often leads to suboptimal behavior, including unnecessary overthinking on simple queries or insufficient reasoning for complex ones. Although some existing approaches allow manual control over the reasoning budget, they still rely on explicit human intervention and cannot adaptively adjust their depth based on the task, limiting their practical usefulness. To address these limitations, we design an adaptive hybrid reasoning mechanism that enables the model to dynamically adjust the depth of reasoning according to the semantic requirements of each question.

Specifically, the model naturally adjusts its reasoning depth based on the information implied in the user’s question as shown in Figure~\ref{plus-fig:cot}(b). Based on the proposed PerCoAct CoT, we categorize user queries into three types. Perception-oriented questions like those asking what is happening or what objects are present naturally call for short reasoning chains grounded primarily in visual observation. Cognition-oriented questions require a deeper analysis of abnormal events, causal factors, norm violations, and potential consequences. Therefore, they activate both the perception and cognition stages. Action-oriented questions, which inquire about risk, potential danger, or recommended solutions, invoke the complete Perception–Cognition–Action reasoning process. This mechanism allows the model to adaptively determine the depth of reasoning based on the user queries, thereby adapting to the diverse needs of real-world applications.

\subsubsection{Video Collection}

Existing video anomaly detection (VAD) datasets generally lack annotations describing the underlying reasoning process, making them unsuitable for the development of video anomaly reasoning (VAR) models. To construct a dataset aligned with the needs of VAR, we consider two key aspects during video collection. First, we aim to cover a diverse range of real-world scenarios. Following the collection strategy of HAWK~\cite{tang2024hawk}, we gather videos from widely used VAD benchmarks, including crime scenes under surveillance (UCF-Crime~\cite{sultani2018real}), violent events captured by cameras (XD-Violence~\cite{wu2020not}), traffic incidents (TAD~\cite{lv2021localizing}), campus environments (ShanghaiTech~\cite{luo2021future}), and urban public scenes (UBnormal~\cite{acsintoae2022ubnormal}). In addition, we incorporate videos from ECVA~\cite{du2024exploring}, a multi-scene benchmark designed to support diverse video analysis tasks. Second, we seek to broaden the coverage of anomaly categories to better support comprehensive reasoning. To this end, we define a hierarchical taxonomy of anomalies consisting of three primary types: Human Activity Anomaly, Environmental Anomaly, and Object-Related Anomaly. As shown in Figure~\ref{plus-fig:dataset-comparison}(a), each type is further divided into several main categories and fine-grained subcategories. Based on this taxonomy, we additionally collect videos from the internet to expand the diversity of anomaly patterns not fully represented in existing datasets. In total, the Vad-Reasoning-Plus dataset contains 8,203 training videos and 438 test videos. To balance annotation cost with the requirement for large-scale training, the dataset is organized into two complementary subsets: Vad-Reasoning-Plus-SFT, consisting of 1,755 videos with high-quality PerCoAct reasoning annotations, and Vad-Reasoning-Plus-RL, containing 6,448 videos annotated only with video-level weak labels.

\subsubsection{Question Annotation}

To construct the proposed Vad-Reasoning-Plus dataset, we first manually annotate the fine-grained anomaly categories of the video, the duration of the anomaly events and the risk level corresponding to anomaly categories. Then we annotate the dataset based on two proprietary models Qwen-Max~\cite{yang2025qwen3} and Qwen-VL-Max~\cite{Qwen3vl}. To further ensure the correctness of the generated CoT, all annotations are firstly verified by large language models and subsequently checked by human experts.

To support multi-level adaptive video anomaly reasoning, each video is paired with a set of questions designed to probe different depths of reasoning. We construct a library of 37 question templates spanning three reasoning dimensions aligned with the PerCoAct CoT(detailed in Section~\ref{plus-sec:cot}). The perception templates focus on visual scene interpretation, object identification, and motion cues. The cognition templates address anomaly categorization, causal factors, intentions, norm violations, and consequence prediction. The action templates target risk evaluation, response planning, and preventive strategies. Each template is associated with the reasoning stages it may reference, ensuring that the question structure is consistent with the intended reasoning depth. Please refer to Appendix C for more details about the question templates.

Using this template library, we construct six questions per video, consisting of three multiple-choice and three open-ended questions, with one question drawn from each reasoning depth. This guarantees that every video is evaluated across perception, cognition, and action level reasoning without redundancy or overlap in question types. The correct answers in multiple-choice questions are randomly positioned to avoid positional bias.

For videos in Vad-Reasoning-Plus-SFT subset, each question is further accompanied by a detailed reasoning process. We first generate a corresponding set of questions based on the video content, and then annotate each question with structured PerCoAct CoT as detailed in Section~\ref{plus-subsubsec:reasoning_annotation}. For videos in Vad-Reasoning-Plus-RL subset, where full reasoning traces are unnecessary, we instead generate question–answer pairs using template-guided prompts based on a concise video description. These weakly annotated QA pairs provide sufficient supervision for RL while enabling large-scale data coverage.

\begin{table*}[t]
\centering
\caption{Dataset comparison with detailed annotation and reasoning fields.
$^{\dagger}$ indicates our preliminary version.}
{
\footnotesize
\setlength{\tabcolsep}{2pt}
\begin{tabular}{l c c c c c c c c c c c c}
\toprule
\multirow{2}{*}{Dataset} &
\multirow{2}{*}{Videos} &
\multirow{2}{*}{Duration} &
\multirow{2}{*}{Samples} &
\multirow{2}{*}{Anomalies} &
\multicolumn{5}{c}{Annotation} &
\multicolumn{3}{c}{Reasoning} \\
\cmidrule(lr){6-10}\cmidrule(lr){11-13}
 & & & & &
Category & Description & Questions & Risk & Action &
Type & Structured & Hierarchy \\
\midrule

\rowcolor{gray!15}
\multicolumn{13}{c}{\textit{Traditional Video Anomaly Detection Datasets}} \\
UCF-Crime~\cite{sultani2018real}              & 1900 & 128h   & 1900  & 13 & \cmark & \xmark & -- & \xmark & \xmark & --   & \xmark & \xmark \\
XD-Violence~\cite{wu2020not}                  & 4754 & 217h   & 4754  &  6 & \cmark & \xmark & -- & \xmark & \xmark & --   & \xmark & \xmark \\
ShanghaiTech~\cite{luo2021future}             &  437 & --     &  437  & 13 & \xmark & \xmark & -- & \xmark & \xmark & --   & \xmark & \xmark \\
UCSD Ped1~\cite{li2013anomaly}                &   70 & 0.1h   &   70  &  5 & \xmark & \xmark & -- & \xmark & \xmark & --   & \xmark & \xmark \\
UCSD Ped2~\cite{li2013anomaly}                &   28 & 0.1h   &   28  &  5 & \xmark & \xmark & -- & \xmark & \xmark & --   & \xmark & \xmark \\
CUHK Avenue~\cite{lu2013abnormal}             &   37 & 0.3h   &   37  &  5 & \xmark & \xmark & -- & \xmark & \xmark & --   & \xmark & \xmark \\
TAD~\cite{lv2021localizing}                   &  518 & 1.2h   &  518  &  7 & \xmark & \xmark & -- & \xmark & \xmark & --   & \xmark & \xmark \\
UBnormal~\cite{acsintoae2022ubnormal}         &  543 & 2.2h   &  543  & 22 & \xmark & \xmark & -- & \xmark & \xmark & --   & \xmark & \xmark \\
NWPU Campus~\cite{cao2024scene}                &  547 & 16.3h  &  547  & 28 & \xmark & \xmark & -- & \xmark & \xmark & --   & \xmark & \xmark \\

\midrule
\rowcolor{gray!15}
\multicolumn{13}{c}{\textit{Video Anomaly Understanding Datasets and Benchmarks}} \\
UCA~\cite{yuan2024towards}                    & 1854 & 121.9h & 1854   & 13 & \cmark & \cmark & --  & \xmark & \xmark & --    & \xmark & \xmark \\
CUVA~\cite{du2024uncovering}                  &  986 & 32.5h  & 5030   & 42 & \cmark & \cmark & 5  & \xmark & \xmark & task  & \xmark & \xmark \\
ECVA~\cite{du2024exploring}                   & 2240 & 88.2h  & 6720   & 100& \cmark & \cmark & 5  & \xmark & \xmark & task  & \xmark & \xmark \\
HIVAU-70k~\cite{zhang2024holmesvau}           & 5593 & 266.5h & 13946  & 13 & \xmark & \cmark & 4  & \xmark & \xmark & task  & \xmark & \xmark \\
HAWK~\cite{tang2024hawk}                      & 7898 & 142.5h & 15196  & -- & \xmark & \cmark & 7  & \xmark & \xmark & --    & \xmark & \xmark \\
VANE~\cite{gani2025vane}                      &  325 & --     &  559   & -- & \xmark & \xmark & 8  & \xmark & \xmark & --    & \xmark & \xmark \\
SurveillanceVQA-589K~\cite{liu2025survqa}     & 3030 & 159.2h & --     & 18 & \cmark & \cmark & 12 & \xmark & \xmark & task  & \xmark & \xmark \\
SmartHome-Bench~\cite{zhao2025smarthome}      & 1203 & --     & 1203   & 29 & \cmark & \cmark & --  & \xmark & \xmark & task  & \xmark & \xmark \\
VRU-Accident~\cite{kim2025vru}                & 1000 & --     & 6000   & -- & \xmark & \cmark & --  & \xmark & \cmark & task  & \xmark & \xmark \\
VAGU~\cite{gao2025vagu}                       & 7567 & --     & \textasciitilde20000 & 21 & \cmark & \cmark & 3 & \xmark & \xmark & task & \xmark & \xmark \\

\midrule
\rowcolor{gray!15}
\multicolumn{13}{c}{\textit{Video Anomaly Reasoning Datasets}} \\
VAU-Bench~\cite{zhu2025vau}                   & 4596 & 169.1h & 4596   & 19 & \cmark & \cmark & 4  & \xmark & \xmark & cot   & \xmark & \xmark \\
Cue-Bench~\cite{yu2025cuebench}               & 2950 & 54.5h  & --     & 32 & \cmark & \cmark & 8  & \xmark & \xmark & cot   & \xmark & \xmark \\
A2Seek~\cite{mo2025a2seek}                    &  542 & --     & --     & 20 & \cmark & \cmark & 4  & \xmark & \xmark & cot   & \cmark & \xmark \\

\midrule
Vad-Reasoning$^{\dagger}$~\cite{huang2025vadr1} & 8641 & 360.5h &  8461 & 37 & \cmark & \cmark & 1  & \xmark & \xmark & cot   & \cmark & \xmark \\
\textbf{Vad-Reasoning-Plus}                     & \textbf{8641} & \textbf{360.5h} & \textbf{50766} & 37 & \cmark & \cmark & \textbf{37} & \cmark & \cmark & cot & \cmark & \cmark \\

\bottomrule
\end{tabular}
}
\label{plus-tab:dataset-comparison}
\end{table*}

\subsubsection{Reasoning Annotation}\label{plus-subsubsec:reasoning_annotation}

In order to ensure that the CoT annotation covers all key information in the video, we follow the principle of high frame information density~\cite{yu2025unhackable}. Specifically, the video is decomposed into separate frames with a frame interval of 16. The extracted frames are then fed into Qwen-VL-Max~\cite{Qwen3vl} to generate detailed dense frame descriptions. Considering the redundancy and high density of video information, directly prompting the model to generate annotations for the entire video would result in high annotation costs and information loss. In comparison, we first perform dense sampling at certain intervals, and then require Qwen-VL-Max~\cite{Qwen3vl} to describe the video frames in detail, which can greatly preserve the key information in the video frames and reduce information loss.

Based on these dense descriptions, we construct the PerCoAct CoT in three steps. We begin by generating the perception stage, which summarizes global scene attributes, object configurations, environmental context, and salient visual dynamics across the video. Next, we produce the cognition stage, where the model identifies abnormal patterns, explains causal factors, infers intentions, detects norm violations, and predicts potential consequences. Finally, we generate the action stage, which includes risk-level estimation, justification for the assigned risk level, suggested immediate responses, and preventive or long-term safety recommendations. Each stage is produced independently through stage-specific prompts, ensuring that the reasoning remains focused, interpretable, and faithfully aligned with the intended semantics. To maintain logical coherence, outputs from earlier stages are provided as inputs to subsequent stages, guaranteeing that the reasoning evolves in a consistent and causally grounded manner.

\subsubsection{Statistical Analysis and Comparison}

We compare Vad-Reasoning-Plus with existing video anomaly detection, understanding, and reasoning datasets in Table~\ref{plus-tab:dataset-comparison}. Vad-Reasoning-Plus consists of 8,641 videos, covering over 360 hours of footage and more than 34 million frames, making it one of the largest datasets for video anomaly tasks. Meanwhile, Vad-Reasoning-Plus dataset is specifically designed for the task of video anomaly reasoning, rather than anomaly detection or isolated anomaly understanding. Existing video anomaly understanding datasets, such as CUVA~\cite{du2024uncovering} and ECVA~\cite{du2024exploring}, primarily support task-level reasoning, where only a limited subset of questions involves causal or consequence-related reasoning. However, their annotations remain isolated and lack an explicit, structured Chain-of-Thought, making it difficult to model a coherent reasoning process. More recent reasoning-oriented datasets like VAU-Bench~\cite{zhu2025vau}, Cue-Bench~\cite{yu2025cuebench}, and A2Seek~\cite{mo2025a2seek} introduce Chain-of-Thought supervision, but their reasoning annotations are unstructured and flat, without an explicit decomposition. As a result, these datasets do not support hierarchical reasoning or stage-aware reasoning control. In contrast, Vad-Reasoning-Plus provides structured and hierarchical Chain-of-Thought annotations, explicitly aligned with a Perception–Cognition–Action paradigm, enabling models to learn stage-aware and progressively deeper reasoning behaviors. Moreover, our dataset offers the largest question coverage, with 37 distinct question types spanning perception, cognition, and action, while additionally incorporating risk assessment and action-oriented annotations, which are absent in prior datasets. This design allows Vad-Reasoning-Plus to uniquely support comprehensive, structured, and hierarchical video anomaly reasoning.

\begin{figure*}[t]
    \centering

    \begin{subfigure}[t]{0.32\textwidth}
        \centering
        \includegraphics[width=\linewidth]{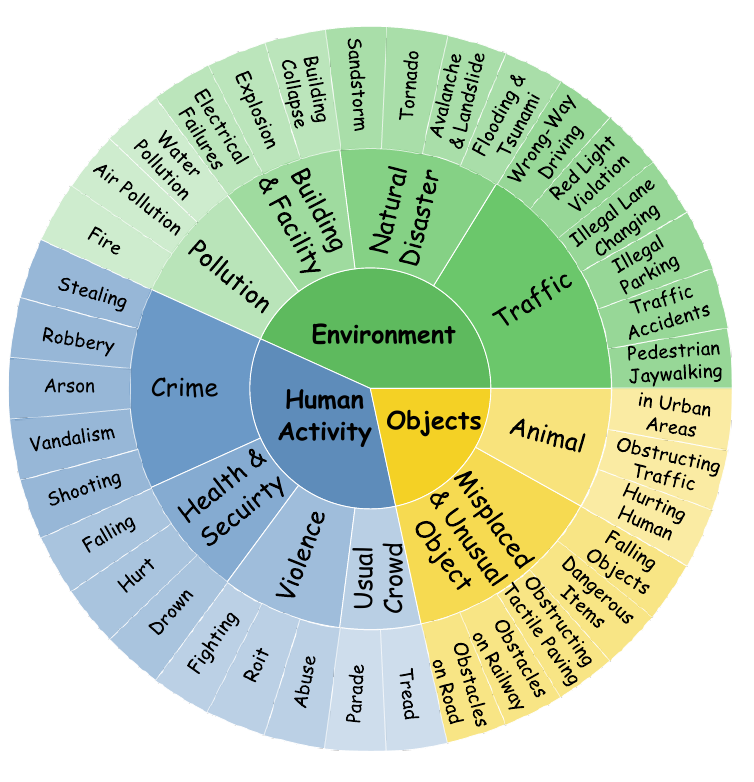}
        \caption{Anomaly categories.}
    \end{subfigure}
    \hfill
    \begin{subfigure}[t]{0.66\textwidth}
        \centering
        \includegraphics[width=\linewidth]{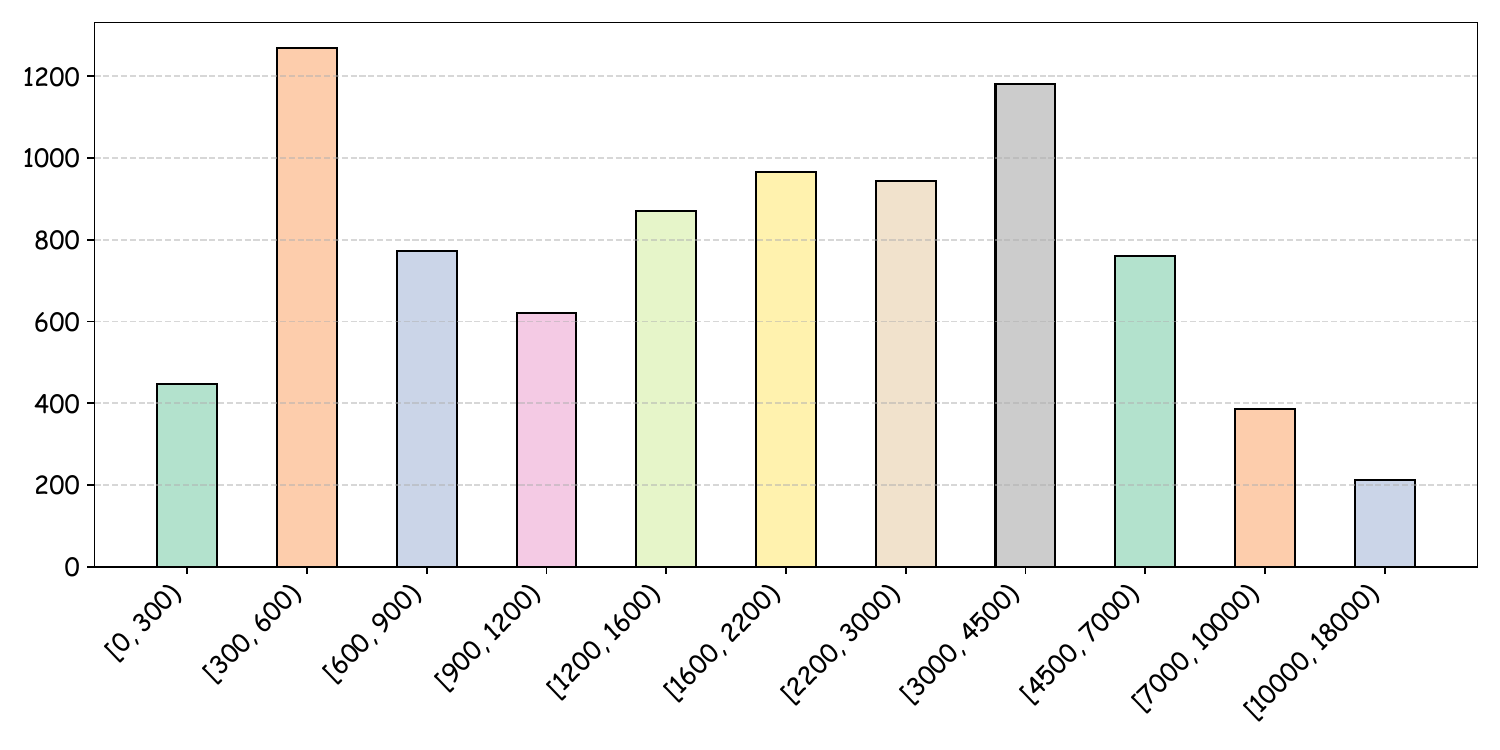}
        \caption{Distribution of video length.}
    \end{subfigure}

    \begin{subfigure}[t]{0.32\textwidth}
        \centering
        \includegraphics[width=\linewidth]{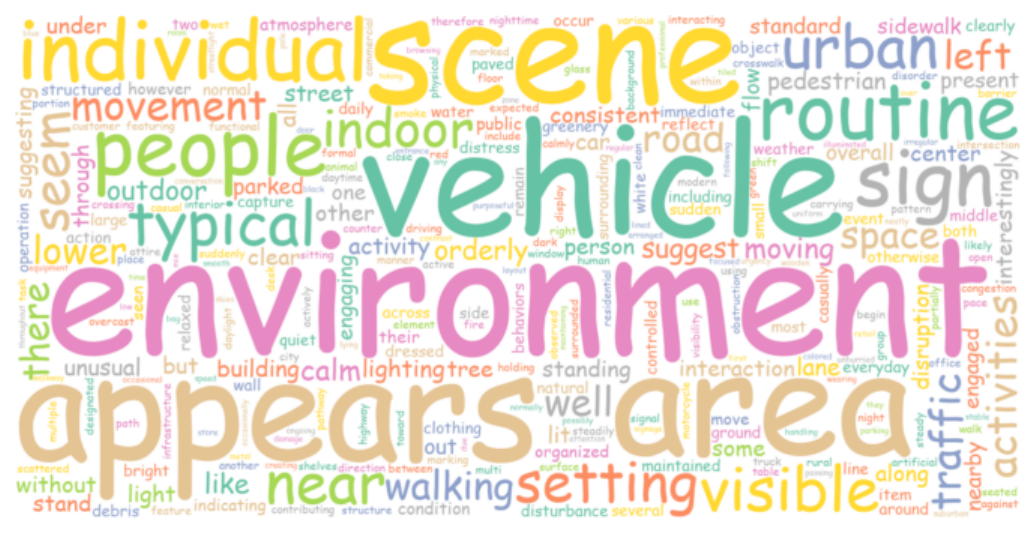}
        \caption{Word cloud of perception stages.}
    \end{subfigure}
    \hfill
    \begin{subfigure}[t]{0.32\textwidth}
        \centering
        \includegraphics[width=\linewidth]{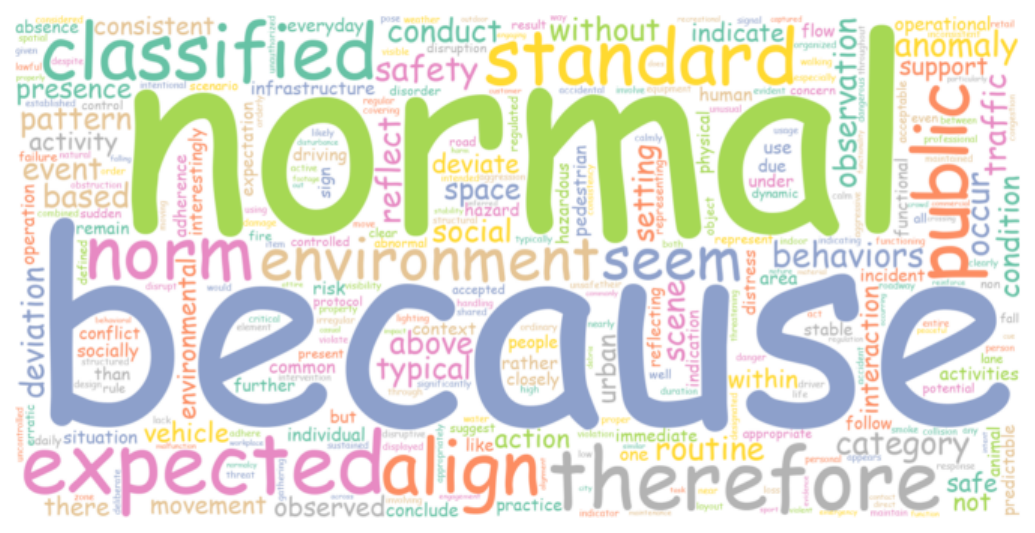}
        \caption{Word cloud of cognition stages.}
    \end{subfigure}
    \hfill
    \begin{subfigure}[t]{0.32\textwidth}
        \centering
        \includegraphics[width=\linewidth]{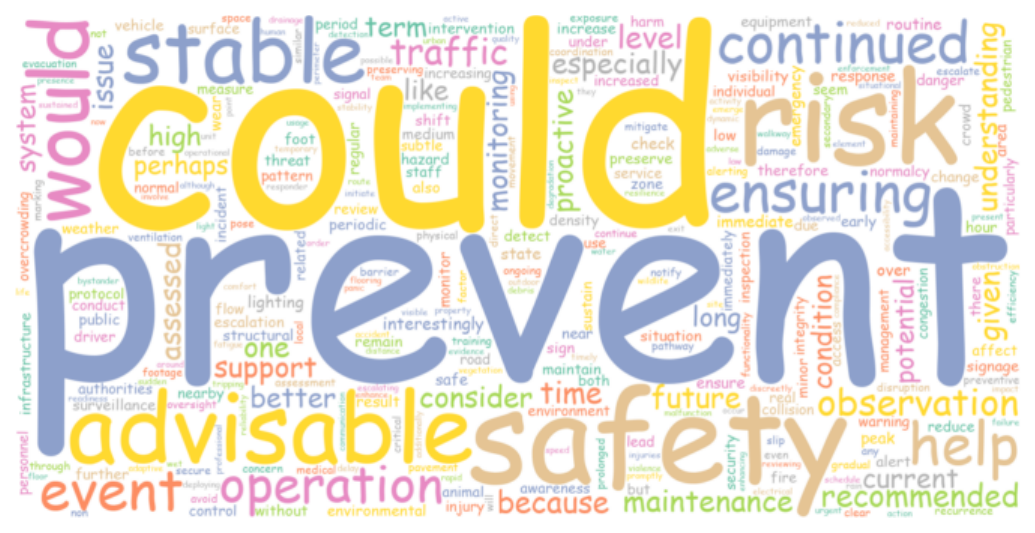}
        \caption{Word cloud of action stages.}
    \end{subfigure}

    \caption{Statistical Analysis of the Vad-Reasoning-Plus Dataset}
    \label{plus-fig:dataset-comparison}
\end{figure*}

Figure~\ref{plus-fig:dataset-comparison} summarizes the statistical characteristics of Vad-Reasoning-Plus. As shown in Figure~\ref{plus-fig:dataset-comparison}(a), the dataset covers diverse anomaly categories across human activities, environments, and objects, reflecting the complexity of real-world surveillance scenarios. Figure~\ref{plus-fig:dataset-comparison}(b) illustrates the distribution of video lengths, which spans from short clips to long sequences, enabling the evaluation of models under varied temporal contexts. Figure~\ref{plus-fig:dataset-comparison}(c)–(e) analyze the linguistic patterns of annotations at different reasoning stages. The perception stage primarily focuses on scene- and object-level observations, the cognition stage emphasizes causal and explanatory semantics, while the action stage highlights risk awareness and decision-oriented expressions. This stage-wise semantic shift demonstrates that Vad-Reasoning-Plus explicitly aligns visual perception, anomaly understanding, and actionable decision making within a unified reasoning framework. Please refer to Appendix B for more details.

\subsection{Perception-Cognition-Action Chain-of-Thought}\label{plus-sec:cot}

Video anomaly reasoning is a domain-specific task, inherently relying on a human-aligned progression from observing the environment to understanding irregular events and ultimately deciding how to respond. To embed such task-specific prior knowledge into models, we design a structured PerCoAct CoT consisting of three stages as shown in Figure~\ref{plus-fig:cot}(a), which formalizes this progression and provides a structured template for comprehensive, interpretable, and decision-oriented reasoning.

\subsubsection{Perception} 

Perception stage focuses on grounding the entire reasoning process in visual evidence by requiring the model to perform holistic observation of the entire video, including environmental conditions, scene layout and objects. In the meanwhile, the model is expected to simultaneously pay attention to suspicious clips where abnormal motion patterns or unusual interactions may occur. Through the combination of global context understanding and local inspection of potentially abnormal segments, this stage ensures that subsequent reasoning is anchored in what is actually visible, instead of inferred from unsupported assumptions.

\subsubsection{Cognition}

Cognition stage performs multi-level semantic interpretation over the observed cues through first forming a judgment of the anomalous phenomenon, describing how the event deviates from normality and explaining the logic behind such deviation. Then the model proceeds toward a deeper understanding of the underlying causes, including violations of norms or unsafe intentions. Furthermore, this stage also considers the possible consequences of the abnormal event, thereby transforming low-level perceptual evidence into a coherent, causally grounded explanation of what the anomaly is and why it matters.

\subsubsection{Action}

Action stage first assesses the severity of the detected situation by estimating a fine-grained risk level and justifying this assessment with reference to the visual and causal cues identified in previous stages. Based on this evaluation, the model generates safety-oriented recommendations, including immediate responses for mitigating the hazard, practical solutions for resolving the current situation, and preventive advice aimed at reducing the likelihood of similar anomalies in the future. This stage aligns closely with real-world requirements, as effective anomaly reasoning must ultimately yield actionable guidance that assists decision-making in safety-critical scenarios.

\subsection{Anomaly-Aware Group Relative Policy Optimization}\label{plus-sec:grpo}

\begin{figure*}[t]
    \centering
    \includegraphics[width=1\linewidth]{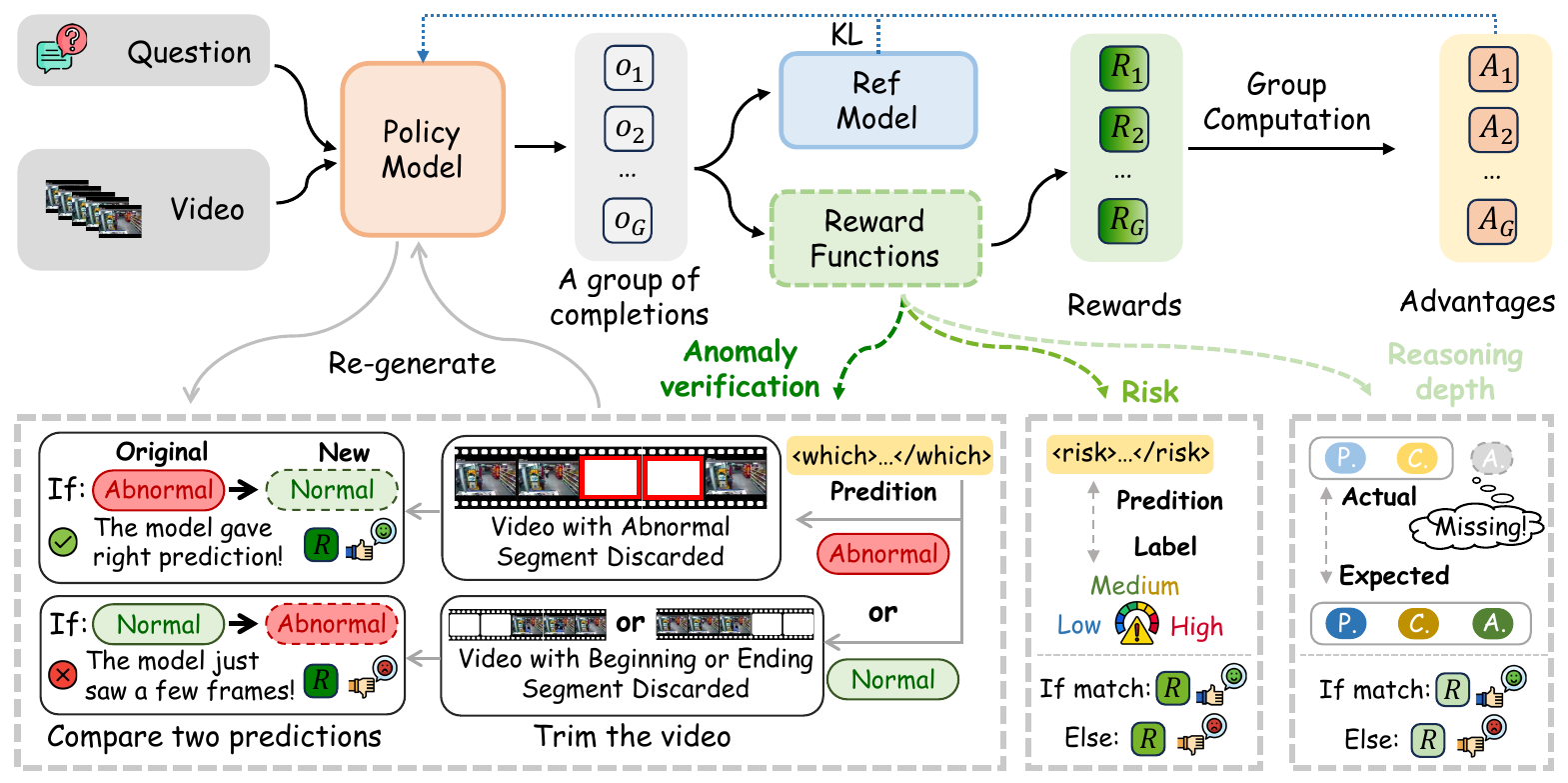} 
    \caption{Illustration of the proposed A\textsuperscript{2}-GRPO.}
    \label{plus-fig:grpo}
\end{figure*}

Since GRPO~\cite{shao2024deepseekmath} is a rule-based reinforcement learning framework, its optimization relies on predefined reward functions. In addition to standard \textbf{Format} and \textbf{Accuracy} rewards, we design a set of anomaly-aware rewards tailored for video anomaly reasoning. These rewards explicitly guide the model toward appropriate reasoning depth, risk-sensitive judgment, and evidence-dependent anomaly judgments, without requiring any additional manual annotations as illustrated in Figure~\ref{plus-fig:grpo}.

\subsubsection{Reasoning Depth Reward}

To regulate the depth of reasoning in a task-aware manner, we design a reasoning-depth reward based on the Perception–Cognition–Action reasoning stages. Given a model response, we identify which reasoning stages are present in the generated output and estimate its effective reasoning depth. For each question, an expected reasoning depth is defined according to its semantic requirements. The reward is then assigned by comparing the predicted reasoning depth with the expected one: responses that include all required stages receive the highest reward, while missing or unnecessary reasoning stages result in progressively lower rewards. Larger mismatches between the predicted and expected depths incur stronger penalties, and invalid or ill-formatted responses receive no reward.

By explicitly measuring whether the model under-reasons or over-reasons with respect to the task, this reward encourages the model to produce reasoning chains that are neither insufficient nor redundant. As a result, the model learns to adapt its reasoning depth to different question types, improving both the efficiency and appropriateness of its reasoning behavior in practical scenarios.

\subsubsection{Risk Reward}

To encourage accurate and safety-aware risk assessment, we design a risk-aware reward based on the alignment between the predicted risk level and the expected risk associated with each event. We define risk with three ordered levels—Low, Medium, and High—reflecting the severity and potential impact of the anomalous event. For each response, the model’s predicted risk level is compared with the expected one, and the reward is assigned according to their discrepancy: exact matches receive the highest reward, while larger deviations result in progressively lower rewards. Predictions that significantly deviate from the expected risk level incur stronger penalties.

By explicitly modeling the distance between predicted and expected risk levels, this reward guides the model toward calibrated and reliable risk judgments. It discourages both underestimation and overestimation of risk, while placing particular emphasis on avoiding severe misjudgments that may lead to unsafe decisions. As a result, the model learns to produce risk assessments that are better aligned with real-world safety requirements and decision-making needs in video anomaly understanding.

\subsubsection{Anomaly Verification Reward}

To assess whether the model’s anomaly judgment is genuinely grounded in abnormal visual evidence rather than spurious cues, we introduce an anomaly verification reward based on temporal evidence manipulation. Given a generated response, we first extract the model’s prediction on anomaly category, and the specific temporal region of the video will be discarded to form a new video clip. We then feed the new video to the model and compare the two judgments from the model.

Specifically, if the video is initially predicted as abnormal, the predicted abnormal segment is temporally discarded to form a trimmed video containing only normal content. The trimmed video is then re-fed into the model. If the regenerated prediction switches from abnormal to normal, it indicates that the original judgment was causally dependent on the abnormal segment, and a positive verification reward is assigned. 

In contrast, we consider the phenomenon of \textbf{\textit{temporal hacking}}~\cite{yu2025unhackable} for video-MLLMs, where models tend to generate predictions by relying on only a few frames, typically the beginning or ending of the video. In this case, if the video is initially predicted as normal, we randomly discard either the beginning or the ending segment of the video and re-evaluate the trimmed input. If the regenerated prediction changes from normal to abnormal, it suggests that the original judgment relied on insufficient evidence. In this case, a negative verification reward is applied to penalize such unreliable reasoning behavior.

By explicitly verifying the stability of anomaly judgments under visual evidence removal, the anomaly verification reward encourages the model to base its predictions on complete and relevant temporal evidence. As a result, the model is guided toward producing anomaly judgments that are more robust, evidence-dependent, and resistant to temporal bias, which is crucial for reliable video anomaly reasoning in real-world scenarios.

\subsubsection{Optimization Objective}

For a question $q$, the model will first generate a group of completions $O = \{o_i\}_{i=0} ^G$. Subsequently, a set of rewards $R = \{r_i\}_{i=0} ^G$ is computed based on the reward functions as described in Section~\ref{plus-sec:grpo}. The rewards are then normalized to compute the relative advantages as

\begin{equation}
    A_i = \frac{r_i - \mathrm{mean}(R)}{\mathrm{std}(R)},
\end{equation}
where $A_i$ is the advantage score of $o_i$, which provides more effective assessment of both individual answer quality and relative comparisons within the group. What's more, to prevent the current policy $\pi_\theta$ from drifting excessively from the reference one $\pi_{\text{ref}}$, GRPO introduces a KL-divergence regularization term. Finally, the objective function of GRPO is formulated as

\begin{equation}
\begin{aligned}
\mathcal{L} = {} & 
\mathbb{E}_{\{q, O\}} \Bigg[ \frac{1}{G} \sum_{i=1}^{G} \Big( 
\min \left( \frac{\pi_\theta(o_t \mid q)}{\pi_{\theta_{\text{old}}}(o_t \mid q)} A_i, \right. \\
& \quad \left. \operatorname{clip}\left( \frac{\pi_\theta}{\pi_{\theta_{\text{old}}}}, 1 - \epsilon, 1 + \epsilon \right) A_i \right) \\
& - \beta\, \mathrm{D_{KL}} \left( \pi_\theta \parallel \pi_{n\theta} \right) \Big) \Bigg],
\label{plus-equ:grpo}
\end{aligned}
\end{equation}
where $\frac{\pi_\theta(o_i \mid q)}{\pi_{\theta_{\text{old}}}(o_i \mid q)}$ quantifies the relative change between the current policy and the old one, and the $\operatorname{clip}\left( \cdot, 1 - \epsilon, 1 + \epsilon \right)$ operation constrains the ratio within a range.

\subsubsection{Training Pipeline}

To balance annotation cost and reasoning capability, we adopt a two-phase training pipeline. In the first stage, supervised fine-tuning(SFT) is conducted on the Vad-Reasoning-Plus-SFT dataset, where videos are annotated with high-quality, structured Chain-of-Thought following the proposed Perception–Cognition–Action paradigm. In this stage, the model is gradually adapted from general multimodal understanding to video anomaly reasoning, learning to produce coherent, stage-aware reasoning processes aligned with anomaly tasks. The training objective of the SFT stage is formulated as 

\begin{equation}
    L=-\sum_{t=1}^T \log P(y_t \mid y_{<t}, x),
\end{equation}
where $y_t$ denotes the ground-truth token at time step $t$, $y_{<t}$ represents the preceding tokens, and $x$ is the multimodal input. This stage primarily establishes the model’s ability to generate structured reasoning outputs and follow the predefined reasoning format.

In the second stage, the model is further optimized on the Vad-Reasoning-Plus-RL dataset using the proposed A\textsuperscript{2}-GRPO. Unlike the SFT stage, the RL dataset does not contain explicit Chain-of-Thought annotations. The model generates structured responses autonomously, which are then evaluated by the anomaly-aware reward functions introduced in Section~\ref{plus-sec:grpo}. This design avoids the need for additional manual reasoning annotations while encouraging the model to refine its reasoning depth, risk assessment, and anomaly judgments through self-verification.

\section{Experiments}
\subsection{Experimental Settings}
\subsubsection{Implementation Details}

Vad-R1-Plus is trained in a two-stage manner based on Qwen2.5-VL-7B~\cite{bai2025qwen25vl}. In the supervised fine-tuning (SFT) stage, the base model is trained on the Vad-Reasoning-Plus-SFT dataset for one epoch.
In the reinforcement learning stage, training is continued on the Vad-Reasoning-Plus-RL dataset using the proposed A\textsuperscript{2}-GRPO. Due to computational constraints, the RL stage is conducted for 1000 optimization steps unless otherwise specified. For efficiency, we uniformly sample 16 frames from the video, and the maximum number of pixels per frame is limited to $128 \times 28 \times 28$ during training. During evaluation, the same preprocessing configurations are applied unless otherwise specified. The number of completions generated in a group is set to 4.

During both training and evaluation, the model is prompted to generate responses in a consistent structured format consisting of a reasoning trace enclosed by \texttt{<think>} tags and a final answer enclosed by \texttt{<answer>} tags. Within the reasoning trace, the model adaptively generates up to three ordered stages (\texttt{<perception>}, \texttt{<cognition>}, and \texttt{<action>}) according to the type and complexity of the question, omitting unnecessary stages instead of enforcing a fixed-depth reasoning chain. At each reasoning stage, the model is required to produce specific structured information that is explicitly encoded using predefined tags. For cognition-stage outputs, the model explicitly predicts the anomaly or activity category using \texttt{<which>} tags, and for abnormal videos, the temporal interval of the anomaly using normalized timestamps enclosed in \texttt{<when>} tags. For action-stage outputs, the model further assesses event severity by predicting a risk level using \texttt{<risk>} tags when applicable. This structured output design enables reliable parsing of reasoning depth, temporal localization, and risk assessment for both reinforcement learning and evaluation. Please refer to Appendix A for more implementation details.

\begin{table*}[t]
\centering
\caption{Evaluation of reasoning on Vad-Reasoning-Plus. $^{*}$ Qwen3-VL~\cite{Qwen3vl} and Kimi-VL~\cite{team2025kimi} appear twice in the table as they represent different variants within the same model family, including instruction-tuned and thinking-enabled versions, which share the same base architecture but exhibit distinct capabilities.}

{
\footnotesize
\setlength{\tabcolsep}{1pt}
\begin{tabular}{lc|ccc|cccc|ccc}
\toprule
\multirow{2}{*}{\textbf{Method}} & 
\multirow{2}{*}{\textbf{Para.}} & 
\multicolumn{3}{c|}{\textbf{Lexical}} & 
\multicolumn{4}{c|}{\textbf{Semantic}} & 
\multicolumn{3}{c}{\textbf{Judgment-based}} \\
\cmidrule(lr){3-5} \cmidrule(lr){6-9} \cmidrule(lr){10-12}
 &  & 
\textbf{BLEU} & \textbf{METEOR} & \textbf{ROUGE} & 
\textbf{SentenceBERT} & \textbf{Qwen3} & \textbf{BLEURT} & \textbf{BERTScore} & 
\textbf{Reasonability}  & \textbf{Detail} & \textbf{Consistency} \\
\midrule
\rowcolor{gray!15}
\multicolumn{12}{c}{\textit{Video MLLMs}} \\
InternVideo2.5~\cite{wang2025internvideo2}  & 8B & 0.002 & 0.151 & 0.031 & 0.795 & 0.785 & 0.479 & 0.823 & 0.608 & 0.550 & 0.512 \\
InternVL3.5~\cite{wang2025internvl3}        & 8B & 0.003 & 0.233 & 0.062 & 0.831 & 0.661 & 0.467 & 0.853 & 0.741 & 0.669 & 0.671 \\
VideoChat-Flash~\cite{li2024videochat}      & 7B & 0.000 & 0.035 & 0.017 & 0.721 & 0.687 & 0.338 & 0.836 & 0.328 & 0.212 & 0.283 \\
VideoLLaMA3~\cite{zhang2025videollama}      & 7B & 0.000 & 0.036 & 0.010 & 0.606 & 0.713 & 0.317 & 0.815 & 0.238 & 0.172 & 0.230 \\
LLaVA-NeXT-Video~\cite{zhang2024video}      & 7B & 0.001 & 0.096 & 0.035 & 0.723 & 0.612 & 0.400 & 0.844 & 0.636 & 0.513 & 0.614 \\
LLaVA-OneVision~\cite{li2024llavaone}       & 7B & 0.000 & 0.048 & 0.023 & 0.747 & 0.658 & 0.375 & 0.839 & 0.543 & 0.372 & 0.585 \\
Qwen3VL-Instruct$^{*}$~\cite{Qwen3vl}       & 8B & 0.006 & 0.198 & 0.074 & 0.827 & 0.769 & 0.460 & 0.859 & 0.768 & 0.702 & 0.711 \\
Kimi-VL-Instruct$^{*}$~\cite{team2025kimi}& 16B-A3B & 0.005 & 0.211 & 0.072 & 0.810 & 0.736 & 0.480 & 0.838 & 0.766 & 0.676 & 0.696 \\
Video-XL~\cite{shu2025video}                & 7B & 0.000 & 0.012 & 0.006 & 0.593 & 0.513 & 0.230 & 0.544 & 0.431 & 0.369 & 0.452 \\
\midrule
\rowcolor{gray!15}
\multicolumn{12}{c}{\textit{Video Reasoning MLLMs}} \\
OpenR1-Video~\cite{wang-2025-open-r1-video} & 7B & 0.003 & 0.132 & 0.060 & 0.793 & 0.743 & 0.441 & 0.848 & 0.660 & 0.547 & 0.642 \\
Video-R1~\cite{feng2025video}               & 7B & 0.006 & 0.233 & 0.080 & 0.824 & 0.754 & 0.497 & 0.861 & 0.775 & 0.701 & 0.701 \\
VideoChat-R1~\cite{li2025videochat}         & 7B & 0.004 & 0.162 & 0.073 & 0.840 & 0.767 & 0.453 & 0.861 & 0.756 & 0.646 & 0.706 \\
Qwen3VL-Thinking$^{*}$~\cite{Qwen3vl}       & 8B & 0.007 & 0.251 & 0.082 & 0.788 & 0.789 & 0.412 & 0.855 & 0.751 & 0.710 & 0.695 \\
Keye-VL1.5~\cite{yang2025kwai}              & 8B & 0.003 & 0.224 & 0.051 & 0.804 & 0.685 & 0.461 & 0.836 & 0.761 & 0.682 & 0.667 \\
MiMo-VL~\cite{xiaomi2025mimo}               & 7B & 0.005 & 0.205 & 0.058 & 0.817 & 0.725 & 0.455 & 0.840 & 0.787 & 0.718 & 0.732 \\
MiniCPM-V4.5~\cite{yu2025minicpm}           & 8B & 0.005 & 0.228 & 0.064 & 0.824 & 0.705 & 0.490 & 0.841 & 0.797 & 0.735 & 0.748 \\
Kimi-VL-Thinking$^{*}$~\cite{team2025kimi}& 16B-A3B & 0.004 & 0.221 & 0.061 & 0.786 & 0.686 & 0.447 & 0.843 & 0.799 & 0.744 & 0.716 \\
Ovis2.5~\cite{lu2025ovis2}                  & 9B & 0.003 & 0.191 & 0.048 & 0.789 & 0.707 & 0.446 & 0.823 & 0.775 & 0.703 & 0.723 \\
GLM4.1-V~\cite{v2507glm}                    & 9B & 0.003 & 0.214 & 0.055 & 0.791 & 0.617 & 0.458 & 0.836 & 0.730 & 0.778 & 0.644 \\

\midrule
\rowcolor{gray!15}
\multicolumn{12}{c}{\textit{MLLM-based VAD Methods}} \\
Holmes-VAD~\cite{zhang2024holmes}           & 2B & 0.002 & 0.092 & 0.040 & 0.765 & 0.728 & 0.409 & 0.841 & 0.570 & 0.465 & 0.526 \\
Holmes-VAU~\cite{zhang2024holmesvau}        & 7B & 0.000 & 0.053 & 0.027 & 0.771 & 0.730 & 0.396 & 0.836 & 0.566 & 0.395 & 0.594 \\
HAWK~\cite{tang2024hawk}                    & 7B & 0.002 & 0.180 & 0.052 & 0.782 & 0.620 & 0.455 & 0.843 & 0.557 & 0.556 & 0.435 \\
\midrule
\rowcolor{gray!15}
\multicolumn{12}{c}{\textit{Proprietary MLLMs}} \\
Gemini2.5-Flash~\cite{comanici2025gemini}   &  - & 0.006 & 0.251 & 0.071 & 0.825 & 0.737 & 0.458 & 0.855 & 0.826 & 0.787 & 0.743 \\
GPT5-mini~\cite{gpt5}                       &  - & 0.002 & 0.187 & 0.052 & 0.835 & 0.761 & 0.447 & 0.839 & 0.848 & 0.791 & 0.795 \\
Grok4-Fast~\cite{grok4}                     &  - & 0.004 & 0.243 & 0.072 & 0.820 & 0.781 & 0.490 & 0.861 & 0.852 & 0.819 & 0.778 \\
\midrule
\rowcolor{gray!15}
\multicolumn{12}{c}{\textit{Ours}} \\
Vad-R1~\cite{huang2025vadr1}(Preliminary)    & 7B & 0.009 & 0.236 & 0.096 & 0.861 & 0.816 & 0.505 & 0.870 & 0.806 & 0.712 & 0.757 \\
\textbf{Vad-R1-Plus}      & 7B & \textbf{0.086} & \textbf{0.387} & \textbf{0.211} & \textbf{0.950} & \textbf{0.927} & \textbf{0.561} & \textbf{0.896} &  \textbf{0.876} & \textbf{0.826} & \textbf{0.820}   \\
\bottomrule
\end{tabular}
}
\label{plus-tab:reasoning-evaluation}
\end{table*}

\subsubsection{Evaluation Metrics and Baselines}

We evaluate Vad-R1-Plus on the test set of Vad-Reasoning-Plus, where each video is associated with three types of questions, requiring different reasoning depths. For multiple-choice questions, we report accuracy as the primary metric. For the open-ended questions, we primarily measure surface-form correctness and faithfulness to reference responses with \textbf{lexical-level metrics}, including BLEU~\cite{papineni2002bleu}, ROUGE~\cite{lin2004rouge} and METEOR~\cite{banerjee2005meteor}. Beyond the answers, we explicitly evaluate the quality of the generated intermediate  reasoning process. Compared with short final answers, reasoning outputs are typically longer, more diverse, and require semantic consistency and evidence dependence. Therefore, we adopt a comprehensive evaluation with three levels: (1) \textbf{Lexical-level} metrics to measure surface alignment; (2) \textbf{Semantic-level} metrics to capture contextual meaning similarity beyond word matching, including BERTScore~\cite{zhang2019bertscore}, BLEURT~\cite{sellam2020bleurt} and two embedding models~\cite{zhang2025qwen3-embedding, reimers-2019-sentence-bert}; (3) \textbf{Judgment-level} metrics, where a strong LLM is used as a grader to assess reasoning quality along multiple aspects~\cite{tang2024hawk}. Furthermore, since VAR targets structured and hierarchical reasoning, we conduct a structure-aware analysis based on the generated CoT, including anomaly category accuracy, risk level accuracy for action-oriented questions and reasoning depth alignment with task requirements. Finally, we additionally consider complementary consistency-based and preference-based evaluations. Please refer to Appendix C for more experimental results.

To demonstrate the effectiveness and generality of Vad-R1-Plus for video anomaly reasoning, we compare it against a diverse set of baselines covering different modeling paradigms, including general video MLLMs~\cite{wang2025internvideo2, wang2025internvl3, li2024videochat, zhang2025videollama, li2024llavaone, Qwen3vl, team2025kimi, shu2025video}, reasoning-oriented video MLLMs~\cite{wang-2025-open-r1-video, feng2025video, li2025videochat, Qwen3vl, yang2025kwai, xiaomi2025mimo, yu2025minicpm, team2025kimi, lu2025ovis2, v2507glm}, MLLM-based VAD methods~\cite{zhang2024holmes, zhang2024holmesvau, tang2024hawk} and proprietary models~\cite{comanici2025gemini, gpt5, grok4}. Moreover, we also consider our preliminary Vad-R1~\cite{huang2025vadr1}.

\subsection{Main Results}

In this section, we present the main experimental results of Vad-R1-Plus on the Vad-Reasoning-Plus benchmark. Since video anomaly reasoning requires not only correct final answers but also reliable and interpretable reasoning processes, we conduct a comprehensive evaluation from the perspectives of both reasoning and answer.


\begin{table}[t]
\centering
\small
\setlength{\tabcolsep}{2pt}
\caption{Evaluation of answers on Vad-Reasoning-Plus. MCQ is short for Multi-Choice Questions and OpenQA is short for Open-ended Question-Answering.}

\begin{tabular}{lcccc}
\toprule
\multirow{2}{*}{\textbf{Method}} 
& \multirow{2}{*}{\textbf{MCQ}} 
& \multicolumn{3}{c}{\textbf{OpenQA}} \\
\cmidrule(lr){3-5}
& 
& \textbf{BLEU-3} 
& \textbf{BLEU-4} 
& \textbf{ROUGE-2} \\
\midrule
\rowcolor{gray!15}
\multicolumn{5}{c}{\textit{Video MLLMs}} \\
InternVideo2.5~\cite{wang2025internvideo2}  & 0.411 & 0.002 & 0.001 & 0.010 \\
InternVL3.5~\cite{wang2025internvl3}        & 0.751 & 0.009 & 0.005 & 0.031 \\
VideoChat-Flash~\cite{li2024videochat}      & 0.938 & 0.008 & 0.004 & 0.029 \\
VideoLLaMA3~\cite{zhang2025videollama}      & 0.771 & 0.003 & 0.002 & 0.014 \\
LLaVA-NeXT-Video~\cite{zhang2024video}      & 0.564 & 0.002 & 0.001 & 0.009 \\
LLaVA-OneVision~\cite{li2024llavaone}       & 0.917 & 0.017 & 0.009 & 0.049 \\
Qwen3VL-Instruct~\cite{Qwen3vl}             & 0.906 & 0.012 & 0.006 & 0.036 \\
Kimi-VL-Instruct~\cite{team2025kimi}        & 0.586 & 0.011 & 0.006 & 0.035 \\
Video-XL~\cite{shu2025video}                & 0.756 & 0.004 & 0.002 & 0.013 \\
\midrule
\rowcolor{gray!15}
\multicolumn{5}{c}{\textit{Video Reasoning MLLMs}} \\
OpenR1-Video~\cite{wang-2025-open-r1-video} & 0.873 & 0.011 & 0.006 & 0.039 \\
Video-R1~\cite{feng2025video}               & 0.916 & 0.004 & 0.002 & 0.010 \\
VideoChat-R1~\cite{li2025videochat}         & 0.916 & 0.011 & 0.006 & 0.037 \\
Qwen3VL-Thinking~\cite{Qwen3vl}             & 0.881 & 0.106 & 0.008 & 0.044 \\
Keye-VL1.5~\cite{yang2025kwai}              & 0.729 & 0.004 & 0.002 & 0.022 \\
MiMo-VL~\cite{xiaomi2025mimo}               & 0.737 & 0.004 & 0.002 & 0.019 \\
MiniCPM-V4.5~\cite{yu2025minicpm}           & 0.722 & 0.011 & 0.005 & 0.037 \\
Kimi-VL-Thinking~\cite{team2025kimi}        & 0.844 & 0.013 & 0.007 & 0.038 \\
Ovis2.5~\cite{lu2025ovis2}                  & 0.915 & 0.009 & 0.004 & 0.032 \\
GLM4.1-V~\cite{v2507glm}                    & 0.002 & 0.001 & 0.000 & 0.008 \\
\midrule
\rowcolor{gray!15}
\multicolumn{5}{c}{\textit{MLLM-based VAD Methods}} \\
Holmes-VAD~\cite{zhang2024holmes}           & 0.566 & 0.002 & 0.001 & 0.009 \\
Holmes-VAU~\cite{zhang2024holmesvau}        & 0.655 & 0.014 & 0.008 & 0.045 \\
HAWK~\cite{tang2024hawk}                    & 0.231 & 0.004 & 0.002 & 0.019 \\
\midrule
\rowcolor{gray!15}
\multicolumn{5}{c}{\textit{Proprietary MLLMs}} \\
Gemini2.5-Flash~\cite{comanici2025gemini}   & 0.910 & 0.019 & 0.010 & 0.050 \\
GPT5-mini~\cite{gpt5}                       & 0.889 & 0.005 & 0.002 & 0.026 \\
Grok4-Fast~\cite{grok4}                     & 0.912 & 0.015 & 0.009 & 0.043 \\
\midrule
\rowcolor{gray!15}
\multicolumn{5}{c}{\textit{Ours}} \\
Vad-R1~\cite{huang2025vadr1}(Preliminary)  & 0.929 & 0.023 & 0.013 & 0.058 \\
\textbf{Vad-R1-Plus}       & \textbf{0.964} & \textbf{0.030} & \textbf{0.019} & \textbf{0.078} \\
\bottomrule
\end{tabular}
\label{plus-tab:answer-evaluation}
\end{table}

\subsubsection{Evaluation on Reasoning}\label{plus-subsubsec:reasoning-evaluation}

Table~\ref{plus-tab:reasoning-evaluation} reports the evaluation on the quality of reasoning, where model-generated reasoning traces are assessed from three complementary perspectives: \textbf{lexical}, \textbf{semantic}, and \textbf{judgment-based} evaluation. At the \textbf{lexical} level, all methods achieve relatively low scores, reflecting the inherent diversity of long-form reasoning traces and the need for complementary semantic- and judgment-based evaluation. At the \textbf{semantic} level, reasoning-oriented MLLMs demonstrate stronger overall capability in capturing the meaning and contextual consistency of reasoning explanations. Compared with prior baselines, Vad-R1-Plus achieves the highest semantic scores across all metrics, indicating more accurate and semantically faithful reasoning aligned with the ground-truth references. At the \textbf{judgment-based} level, we directly assess the overall quality, detail completeness, and logical consistency of the reasoning process. Vad-R1-Plus consistently outperforms both proprietary and open-source baselines on all judgment-based metrics. In particular, Vad-R1-Plus surpasses the proprietary models like Gemini2.5-Flash~\cite{comanici2025gemini}, GPT5-mini~\cite{gpt5}, and Grok4-Fast~\cite{grok4} on \textbf{Consistency} with absolute improvements of 0.077, 0.025, and 0.042, respectively. The advantage of Vad-R1-Plus becomes more pronounced under judgment-based evaluation, which directly assesses the overall quality, detail completeness, and logical consistency of the reasoning process. 

\subsubsection{Evaluation on Answers}\label{plus-subsubsec:answer-evaluation}

We further evaluate the quality of the final answers produced by different methods on Vad-Reasoning-Plus, considering both multiple-choice questions (MCQ) and open-ended question answering (OpenQA). As shown in Table~\ref{plus-tab:answer-evaluation}, Vad-R1-Plus consistently outperforms all competing methods across both evaluation settings. For MCQ, Vad-R1-Plus achieves the highest accuracy among both open-source and proprietary MLLMs, indicating a stronger ability to make correct and reliable decisions under discrete answer constraints. For OpenQA, it also shows clear and consistent advantages across all lexical-level metrics, reflecting more precise and semantically grounded answer generation that is well aligned with the ground-truth responses. Overall, these results demonstrate that Vad-R1-Plus improves not only reasoning quality, but also the correctness, robustness, and practical reliability of final answers in anomaly reasoning scenarios.

\subsection{Ablation Studies}

In this section, we conduct a series of ablation studies to systematically examine how different components contribute to improved video anomaly reasoning capability. We first study the effectiveness of different training strategies in Section~\ref{plus-subsubsec:reward-training}. Then we perform fine-grained ablations on the reward design to disentangle the contributions of anomaly verification, reasoning depth, and risk-aware rewards in Section~\ref{plus-subsubsec:reward-ablation}. 

\subsubsection{Effectiveness of Different Training Strategies}\label{plus-subsubsec:reward-training}

We analyze the impact of different training strategies in Table~\ref{plus-tab:ablation-training}. Firstly, supervised fine-tuning(SFT) leads to a substantial performance improvement across almost all metrics. This indicates that high-quality, structured supervision plays a dominant role in enabling the model to acquire task-specific understanding and produce well-formed anomaly reasoning outputs. In particular, SFT effectively aligns the model with the proposed Perception–Cognition–Action reasoning structure and stabilizes the generation of coherent intermediate reasoning steps. Secondly, applying reinforcement learning(RL) alone yields relatively limited gains compared to SFT, and in some cases even degrades answer-level performance. This behavior is expected, as the RL stage operates on weak video-level supervision without access to explicit reasoning annotations. Without a properly initialized reasoning structure, reward-based optimization alone struggles to guide the model toward meaningful and stable reasoning behaviors. Finally, combining SFT and RL consistently achieves the best overall performance. Building upon the strong initialization provided by SFT, the RL stage further refines the reasoning process by encouraging deeper reasoning, risk-aware decision-making, and improved consistency. This two-stage training paradigm allows reinforcement learning to move beyond surface-level pattern matching learned during SFT, leading to more robust and generalizable video anomaly reasoning capability.


\begin{table}[t]
\centering
\caption{Effectiveness of training strategy. R, D and C are short for \textbf{R}easonability, \textbf{D}etail and \textbf{C}onsistency, respectively. Sen-Bert is short for SentenceBert. $^{\dagger}$ represents Qwen2.5-VL.}
{
\footnotesize
\setlength{\tabcolsep}{2.5pt}
\begin{tabular}{l|ccc|cccc}
\toprule
\multirow{3}{*}{\makecell{\textbf{Training}\\\textbf{Strategy}}} 
& \multicolumn{3}{c|}{\textbf{Answer}} 
& \multicolumn{4}{c}{\textbf{Reasoning}} \\
\cmidrule(lr){2-4}\cmidrule(lr){5-8}

& \multirow{2}{*}{\textbf{MCQ}} 
& \multicolumn{2}{c|}{\textbf{OpenQA}} 
& \textbf{Semantic}
& \multicolumn{3}{c}{\textbf{Judgment-based}} \\
\cmidrule(lr){3-4}\cmidrule(lr){5-5}\cmidrule(lr){6-8}
& & \textbf{BLEU-4} & \textbf{METEOR}
& \textbf{Sen-Bert}
& \textbf{R} & \textbf{D} & \textbf{C} \\
\midrule

Base$^{\dagger}$& 0.898 & 0.008 & 0.089 & 0.843 & 0.762 & 0.663 & 0.704 \\
+SFT   & 0.958 & 0.023 & 0.132 & 0.950 & 0.874 & 0.799 & 0.800 \\
+RL    & 0.940 & 0.009 & 0.094 & 0.846 & 0.781 & 0.670 & 0.730 \\
\textbf{+SFT+RL}   & \textbf{0.964} & \textbf{0.019} & \textbf{0.166} & \textbf{0.954} & \textbf{0.876} & \textbf{0.826} & \textbf{0.820} \\
\bottomrule
\end{tabular}
}

\label{plus-tab:ablation-training}
\end{table}

\begin{table*}[t]
\centering
\setlength{\tabcolsep}{3pt}
\caption{Experiments on effectiveness of different rewards.}
\begin{tabular}{ccc|cc c|ccc ccc}
\toprule
\multicolumn{3}{c|}{\textbf{Reward}} &
\multicolumn{3}{c|}{\textbf{Answer}} &
\multicolumn{6}{c}{\textbf{Reasoning}} \\
\cmidrule(lr){1-3}\cmidrule(lr){4-6}\cmidrule(lr){7-12}

\multirow{2}{*}{\makecell{\textbf{Anomaly}\\\textbf{Verification}}} &
\multirow{2}{*}{\textbf{Risk}} &
\multirow{2}{*}{\makecell{\textbf{Reasoning}\\\textbf{Depth}}} &
\multicolumn{2}{c}{\textbf{OpenQA}} &
\multirow{2}{*}{\textbf{MCQ}} &
\multicolumn{3}{c}{\textbf{Semantic}} &
\multicolumn{3}{c}{\textbf{Judgment-based}} \\
\cmidrule(lr){4-5}\cmidrule(lr){7-9}\cmidrule(lr){10-12}

& & &
\textbf{BLEU-3} & \textbf{METEOR} & &
\textbf{SentenceBert} & \textbf{BLEURTScore} & \textbf{Qwen3} &
\textbf{Reasonability } & \textbf{Detail} & \textbf{Consistency} \\
\midrule
          &          &          & 0.071 & 0.120 & 0.941 & 0.916 & 0.531 & 0.862 & 0.849 & 0.746 & 0.725 \\
\midrule
\ding{51} &          &          & 0.107 & 0.125 & 0.957 & 0.934 & 0.535 & 0.876 & 0.857 & 0.755 & 0.773 \\
          &\ding{51} &          & 0.100 & 0.120 & 0.964 & 0.920 & 0.520 & 0.892 & 0.863 & 0.793 & 0.781 \\
          &          &\ding{51} & 0.102 & 0.118 & 0.951 & 0.938 & 0.541 & 0.913 & 0.842 & 0.784 & 0.773 \\
\midrule
\ding{51} &\ding{51} &          & 0.105 & 0.122 & 0.960 & 0.903 & 0.538 & 0.879 & 0.871 & 0.803 & 0.787 \\
\ding{51} &          &\ding{51} & 0.110 & 0.127 & 0.954 & 0.910 & 0.559 & 0.906 & 0.851 & 0.793 & 0.775 \\
          &\ding{51} &\ding{51} & 0.114 & 0.123 & 0.954 & 0.949 & 0.558 & 0.923 & 0.863 & 0.804 & 0.797 \\
\midrule
\ding{51} &\ding{51} &\ding{51} & \textbf{0.125} & \textbf{0.132} & \textbf{0.964} & \textbf{0.950} & \textbf{0.569} & \textbf{0.927} & \textbf{0.876} & \textbf{0.826} & \textbf{0.820} \\
\bottomrule
\end{tabular}
\label{plus-tab:ablation-rewards}
\end{table*}

\begin{figure*}[t]
    \centering
    \begin{subfigure}{1\textwidth}
        \centering
        \includegraphics[width=\linewidth]{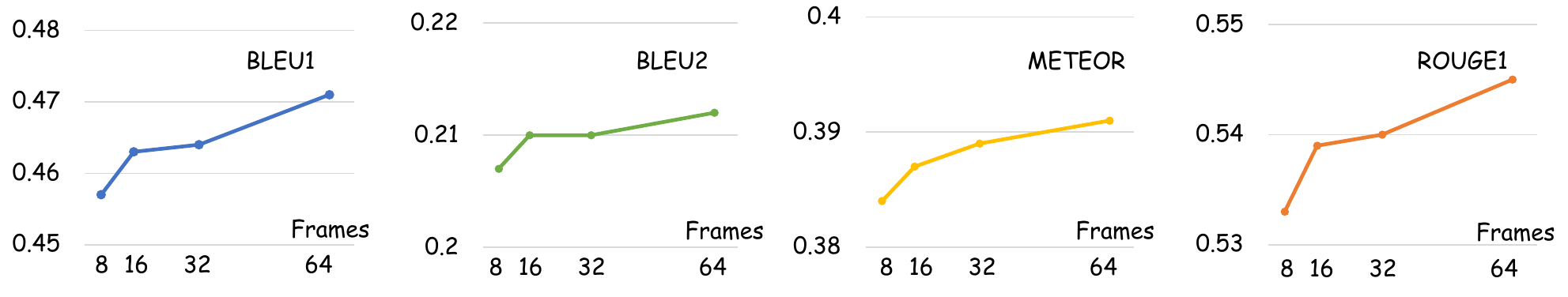}
        \caption{Experiments on the number of input frames.}
    \end{subfigure}
    \begin{subfigure}{1\textwidth}
        \centering
        \includegraphics[width=\linewidth]{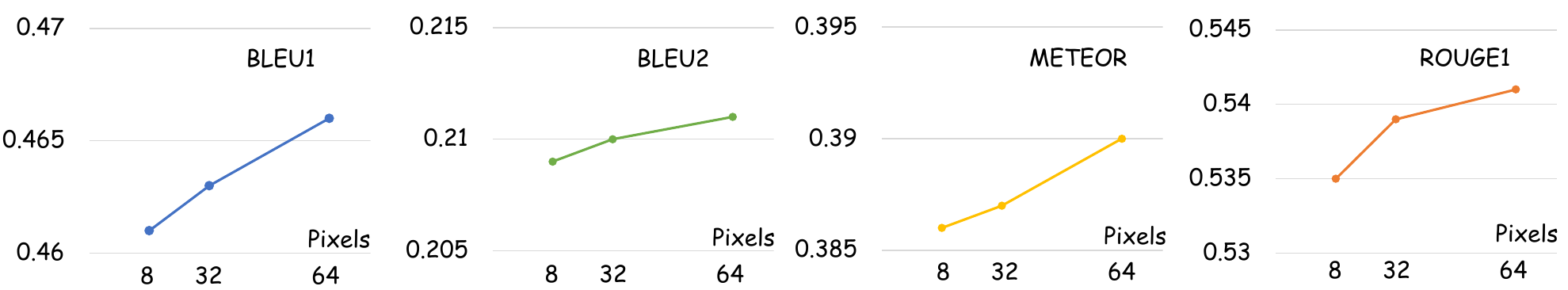}
        \caption{Experiments on the number of input pixels.}
    \end{subfigure}
\caption{Performance comparison on different numbers of input tokens.}
\label{plus-fig:input-tokens}
\end{figure*}

\subsubsection{Effectiveness of the proposed A\textsuperscript{2}-GRPO}\label{plus-subsubsec:reward-ablation}

\begin{table}[t]
\centering
\caption{
Stage-wise evaluation. \textbf{Perception}-oriented questions evaluate the \textbf{Identification} (Ident.) of visual information. \textbf{Cognition}-oriented questions assess anomaly \textbf{Interpretation} (Interp.) and category prediction \textbf{Accuracy} (Acc.). \textbf{Action}-oriented questions focus on the \textbf{Appropriateness} (Approp.) of recommended actions and \textbf{Risk} assessment.
}
{
\footnotesize
\setlength{\tabcolsep}{2pt}
\begin{tabular}{l c cc cc}
\toprule
\multirow{2}{*}{\textbf{Method}} 
& \multicolumn{1}{c}{\textbf{Perception}} 
& \multicolumn{2}{c}{\textbf{Cognition}} 
& \multicolumn{2}{c}{\textbf{Action}} \\
\cmidrule(lr){2-2}\cmidrule(lr){3-4}\cmidrule(lr){5-6}
& \textbf{Ident.} 
& \textbf{Interp.} & \textbf{Acc.} 
& \textbf{Approp.} & \textbf{Risk} \\
\midrule

\rowcolor{gray!15}
\multicolumn{6}{c}{\textit{Video MLLMs}} \\
InternVideo2.5~\cite{wang2025internvideo2}   & 0.684 & 0.684 & 0.519 & 0.631 & 0.469 \\
InternVL3.5~\cite{wang2025internvl3}         & 0.805 & 0.785 & 0.659 & 0.827 & 0.704 \\
VideoChat-Flash~\cite{li2024videochat}       & 0.445 & 0.428 & 0.465 & 0.652 & 0.433 \\
VideoLLaMA3~\cite{zhang2025videollama}       & 0.179 & 0.291 & 0.463 & 0.297 & 0.464 \\
LLaVA-NeXT-Video~\cite{zhang2024video}       & 0.568 & 0.644 & 0.497 & 0.625 & 0.650 \\
LLaVA-OneVision~\cite{li2024llavaone}        & 0.547 & 0.529 & 0.465 & 0.684 & 0.452 \\
Qwen3VL-Instruct~\cite{Qwen3vl}              & 0.803 & 0.829 & 0.860 & 0.822 & 0.765 \\
Kimi-VL-Instruct~\cite{team2025kimi}         & 0.757 & 0.807 & 0.467 & 0.819 & 0.441 \\
Video-XL~\cite{shu2025video}                 & 0.389 & 0.523 & 0.470 & 0.450 & 0.436 \\
\midrule

\rowcolor{gray!15}
\multicolumn{6}{c}{\textit{Video Reasoning MLLMs}} \\
OpenR1-Video~\cite{wang-2025-open-r1-video}  & 0.639 & 0.667 & 0.489 & 0.676 & 0.783 \\
Video-R1~\cite{feng2025video}                & 0.792 & 0.810 & 0.476 & 0.791 & 0.473 \\
VideoChat-R1~\cite{li2025videochat}          & 0.730 & 0.783 & 0.784 & 0.786 & 0.803 \\
Qwen3VL-Thinking~\cite{Qwen3vl}              & 0.779 & 0.816 & 0.802 & 0.824 & 0.753 \\
Keye-VL1.5~\cite{yang2025kwai}               & 0.773 & 0.736 & 0.465 & 0.825 & 0.433 \\
MiMo-VL~\cite{xiaomi2025mimo}                & 0.778 & 0.799 & 0.723 & 0.847 & 0.774 \\
MiniCPM-V4.5~\cite{yu2025minicpm}            & 0.818 & 0.832 & 0.490 & 0.871 & 0.541 \\
Kimi-VL-Thinking~\cite{team2025kimi}         & 0.791 & 0.811 & 0.474 & 0.831 & 0.524 \\
Ovis2.5~\cite{lu2025ovis2}                   & 0.812 & 0.815 & 0.696 & 0.862 & 0.682 \\
GLM4.1-V~\cite{v2507glm}                     & 0.767 & 0.540 & 0.465 & 0.378 & 0.433 \\
\midrule

\rowcolor{gray!15}
\multicolumn{6}{c}{\textit{MLLMs-based VAD Methods}} \\
Holmes-VAD~\cite{zhang2024holmes}            & 0.554 & 0.583 & 0.512 & 0.520 & 0.463 \\
Holmes-VAU~\cite{zhang2024holmesvau}         & 0.573 & 0.580 & 0.465 & 0.625 & 0.433 \\
HAWK~\cite{tang2024hawk}                     & 0.420 & 0.482 & 0.465 & 0.462 & 0.432 \\
\midrule

\rowcolor{gray!15}
\multicolumn{6}{c}{\textit{Proprietary MLLMs}} \\
Gemini2.5-Flash~\cite{comanici2025gemini}    & 0.801 & 0.853 & 0.914 & 0.884 & 0.819 \\
GPT5-mini~\cite{gpt5}                       & 0.827 & 0.862 & 0.856 & 0.918 & 0.850 \\
Grok4-Fast~\cite{grok4}                     & \textbf{0.844} & 0.911 & 0.930 & 0.915 & 0.763 \\
\midrule

\rowcolor{gray!15}
\multicolumn{6}{c}{\textit{Ours}} \\
Vad-R1~\cite{huang2025vadr1}(Preliminary)    & 0.801 & 0.857 & 0.519 & 0.865 & 0.530 \\
\textbf{Vad-R1-Plus}                        & 0.815 & \textbf{0.920} & \textbf{0.966} & \textbf{0.926} & \textbf{0.890} \\
\bottomrule
\end{tabular}
}
\label{plus-tab:reasoning-stage}
\end{table}

We conduct ablation studies to analyze the contribution of each individual reward component in the proposed A\textsuperscript{2}-GRPO. Table~\ref{plus-tab:ablation-rewards} reports the performance under different combinations of anomaly verification, reasoning depth, and risk-aware rewards, evaluated across both answer quality and reasoning quality metrics. Starting from the baseline GRPO~\cite{shao2024deepseekmath} without any anomaly-aware rewards, the model exhibits limited reasoning consistency and suboptimal performance across both answer and reasoning. Introducing the anomaly verification reward leads to noticeable improvements in OpenQA accuracy and semantic reasoning scores, indicating that self-verification through temporal trimming effectively enhances anomaly grounding and reduces spurious predictions. The addition of the reasoning depth reward further improves reasoning-related metrics, particularly reasonability and consistency, demonstrating its effectiveness in encouraging complete and stage-consistent reasoning processes aligned with the Perception–Cognition–Action paradigm. Finally, incorporating the risk-aware reward consistently yields additional gains across both answer and reasoning evaluations. This suggests that explicitly supervising risk assessment not only improves decision-related outputs but also promotes more cautious and coherent reasoning behaviors. When all three reward components are jointly applied, Vad-R1-Plus achieves the best overall performance across OpenQA, MCQ, and multi-level reasoning metrics, further validating the complementary nature of the proposed rewards and their collective effectiveness in enhancing video anomaly reasoning.

\subsection{Discussion}

\subsubsection{Effect of Input Token Quantity}

We investigate the effect of input token quantity on video anomaly reasoning by varying both the number of sampled frames and the spatial resolution of each frame. In the default setting, the model processes 16 frames with a maximum of $128\times28\times28$ pixels per frame, which jointly determines the total number of visual tokens fed into the MLLM. To be more specific, we systematically evaluate configurations with fewer or more visual tokens while keeping all other factors fixed. As illustrated in Figure~\ref{plus-fig:input-tokens}, increasing the amount of visual tokens through denser frame sampling or higher spatial resolution consistently leads to improved anomaly reasoning capabilities. This phenomenon is closely related to the inherent characteristics of video anomaly reasoning, where identifying abnormal events often depends on subtle temporal dynamics and fine-grained spatial cues. When provided with richer visual inputs, the model gains access to more complete temporal context and more detailed spatial context, which helps it better track the progression of anomalous events, detect brief or localized irregular behaviors, and reason about interactions among objects over time. Such enhanced visual evidence further strengthens the perception stage and supports more accurate cognition and more reliable action-level reasoning in turn. Meanwhile, the performance improvements tend to gradually saturate as the number of input tokens continues to increase, indicating that additional visual information eventually becomes redundant. This observation highlights a practical trade-off between computational cost and reasoning effectiveness, where too few tokens risk missing critical evidence, while overly dense inputs provide limited additional benefit. Overall, the results suggest that sufficient visual token coverage is essential for robust video anomaly reasoning. Our default configuration therefore achieves a reasonable balance between efficiency and reasoning quality for practical deployment.

\subsubsection{Stage-wise Evaluation}

To provide a fine-grained analysis of reasoning behavior, we conduct a stage-wise evaluation that aligns question types with their corresponding reasoning requirements, as shown in Table~\ref{plus-tab:reasoning-stage}. Rather than treating reasoning as a monolithic process, this evaluation examines whether models perform the appropriate reasoning for each stage. For \textbf{perception}-oriented questions, we evaluate the \textbf{Identification} of relevant visual information. Most strong video MLLMs achieve competitive performance, indicating that low-level visual perception is generally well handled. Vad-R1-Plus nevertheless attains the highest identification score, suggesting that structured reasoning helps organize perceptual evidence more effectively. For \textbf{cognition}-oriented questions, we assess anomaly \textbf{Interpretation} and category \textbf{Accuracy}. Here, the performance gap becomes more pronounced: many baselines show a clear drop compared to perception, reflecting the challenge of bridging visual observation and semantic abnormality understanding. In contrast, Vad-R1-Plus significantly outperforms all competitors, demonstrating the benefit of explicitly modeling the cognition stage. For \textbf{action}-oriented questions, we focus on response \textbf{Appropriateness} and \textbf{Risk} assessment, which require decision-oriented reasoning beyond recognition and explanation. While most baselines struggle in this setting, Vad-R1-Plus achieves the best performance on both metrics, highlighting the importance of an explicit action stage for risk-aware and practical anomaly reasoning.

\begin{table}[t]
\centering
\setlength{\tabcolsep}{1.5pt}
\caption{Comparison of \textbf{Double Right} and \textbf{Triple Right} metrics. \textbf{Double Right} measures cases where both the reasoning process and the final answer are correct, whereas \textbf{Triple Right} imposes an additional constraint that the anomaly category prediction is also correct, thus requiring full semantic alignment across stages.}
\begin{tabular}{l|cccc|cc}
\toprule
\multirow{2}{*}{\textbf{Method}} 
& \multicolumn{4}{c|}{\textbf{Double Right}} 
& \multicolumn{2}{c}{\textbf{Triple Right}} \\
\cmidrule(lr){2-5}\cmidrule(lr){6-7}
 & \textbf{RR ↑} & \textbf{RW ↓} & \textbf{WR ↓} & \textbf{WW ↓}
 & \textbf{RRR ↑} & \textbf{WWW ↓} \\
\midrule
\rowcolor{gray!15}
\multicolumn{7}{c}{\textit{Video MLLMs}} \\
InternVideo2.5~\cite{wang2025internvideo2} & 0.031 & 0.106 & 0.194 & 0.669 & 0.013 & 0.423 \\
InternVL3.5~\cite{wang2025internvl3}       & 0.069 & 0.138 & 0.361 & 0.432 & 0.015 & 0.330 \\
VideoChat-Flash~\cite{li2024videochat}     & 0.008 & 0.009 & 0.513 & 0.470 & 0.006 & 0.318 \\
VideoLLaMA3~\cite{zhang2025videollama}     & 0.006 & 0.021 & 0.400 & 0.573 & 0.002 & 0.372 \\
LLaVA-NeXT-Video~\cite{zhang2024video}     & 0.009 & 0.025 & 0.279 & 0.687 & 0.002 & 0.465 \\
LLaVA-OneVision~\cite{li2024llavaone}      & 0.007 & 0.009 & 0.526 & 0.459 & 0.005 & 0.311 \\
Qwen3VL-Instruct~\cite{Qwen3vl}            & 0.206 & 0.154 & 0.320 & 0.320 & 0.074 & 0.218 \\
Kimi-VL-Instruct~\cite{team2025kimi}       & 0.172 & 0.221 & 0.190 & 0.417 & 0.042 & 0.256 \\
Video-XL~\cite{shu2025video}               & 0.002 & 0.003 & 0.390 & 0.606 & 0.002 & 0.406 \\
\midrule
\rowcolor{gray!15}
\multicolumn{7}{c}{\textit{Video Reasoning MLLMs}} \\
OpenR1-Video~\cite{wang-2025-open-r1-video}& 0.055 & 0.058 & 0.426 & 0.461 & 0.012 & 0.322 \\
Video-R1~\cite{feng2025video}              & 0.195 & 0.206 & 0.279 & 0.320 & 0.049 & 0.193 \\
VideoChat-R1~\cite{li2025videochat}        & 0.177 & 0.149 & 0.338 & 0.336 & 0.051 & 0.249 \\
Qwen3VL-Thinking~\cite{Qwen3vl}            & 0.247 & 0.197 & 0.279 & 0.276 & 0.087 & 0.194 \\
Keye-VL1.5~\cite{yang2025kwai}             & 0.012 & 0.046 & 0.389 & 0.552 & 0.002 & 0.365 \\
MiMo-VL~\cite{xiaomi2025mimo}              & 0.021 & 0.076 & 0.361 & 0.542 & 0.006 & 0.344 \\
MiniCPM-V4.5~\cite{yu2025minicpm}          & 0.057 & 0.090 & 0.371 & 0.481 & 0.007 & 0.295 \\
Kimi-VL-Thinking~\cite{team2025kimi}       & 0.093 & 0.072 & 0.410 & 0.425 & 0.013 & 0.289 \\
Ovis2.5~\cite{lu2025ovis2}                 & 0.021 & 0.047 & 0.500 & 0.433 & 0.003 & 0.328 \\
GLM4.1-V~\cite{v2507glm}                   & 0.001 & 0.045 & 0.002 & 0.952 & 0.001 & 0.647 \\
\midrule
\rowcolor{gray!15}
\multicolumn{7}{c}{\textit{MLLMs-based VAD Methods}} \\
Holmes-VAD~\cite{zhang2024holmes}          & 0.017 & 0.060 & 0.276 & 0.646 & 0.003 & 0.413 \\
Holmes-VAU~\cite{zhang2024holmesvau}       & 0.007 & 0.029 & 0.365 & 0.600 & 0.006 & 0.397 \\
HAWK~\cite{tang2024hawk}                   & 0.014 & 0.093 & 0.116 & 0.778 & 0.003 & 0.506 \\
\midrule
\rowcolor{gray!15}
\multicolumn{7}{c}{\textit{Proprietary MLLMs}} \\
Gemini2.5-Flash~\cite{comanici2025gemini}  & 0.298 & 0.163 & 0.277 & 0.262 & 0.100 & 0.169 \\
GPT5-mini~\cite{gpt5}                      & 0.219 & 0.145 & 0.286 & 0.350 & 0.077 & 0.190 \\
Grok4-Fast~\cite{grok4}                    & 0.360 & 0.220 & 0.202 & 0.218 & 0.141 & 0.135 \\
\midrule
\rowcolor{gray!15}
\multicolumn{7}{c}{\textit{Ours}} \\
Vad-R1~\cite{huang2025vadr1}(Preliminary)  & 0.286 & 0.160 & 0.292 & 0.263 & 0.076 & 0.164 \\
\textbf{Vad-R1-Plus}           & \textbf{0.571} & 0.252 & 0.087 & 0.090 & \textbf{0.467} & 0.069 \\
\bottomrule
\end{tabular}

\label{plus-tab:double-right}
\end{table}

\subsubsection{Reasoning Depth Analysis}

\begin{figure*}[t]
    \centering
    \includegraphics[width=1\linewidth]{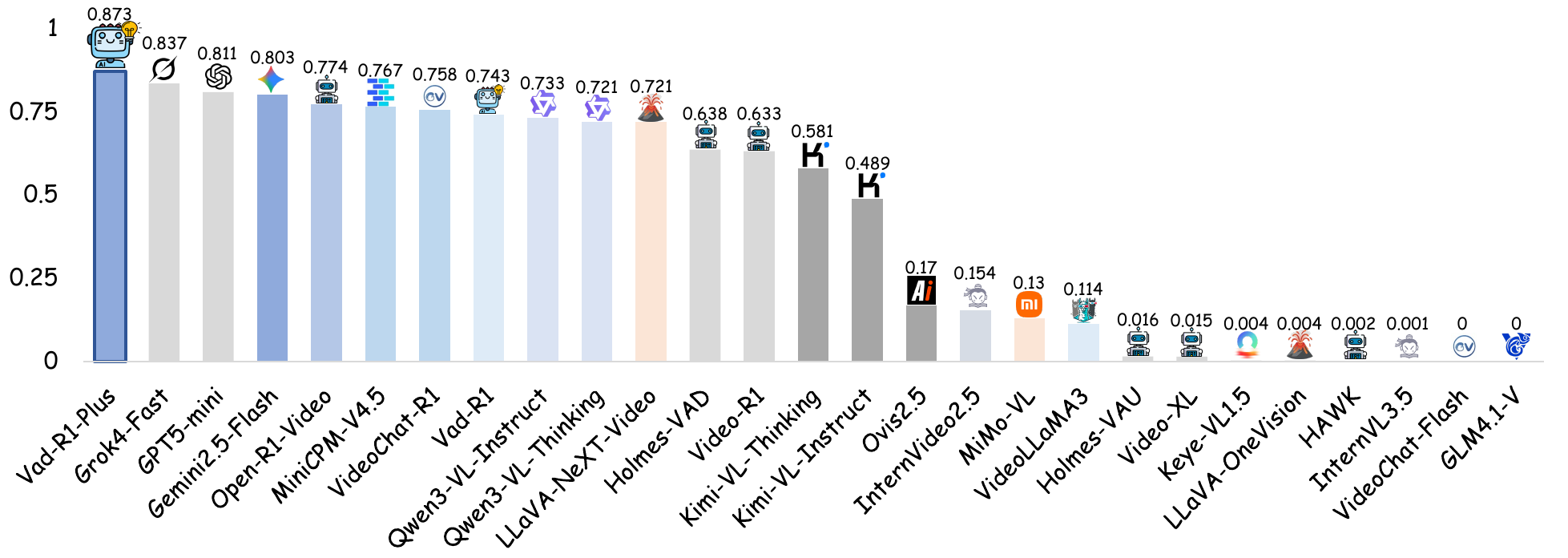} 
    \caption{Comparison of reasoning depth alignment across different models.}
    \label{plus-fig:reasoning-depth}
\end{figure*}

We further analyze the ability of different models to adaptively adjust their reasoning depth according to question requirements. As video anomaly reasoning involves heterogeneous queries spanning perception, cognition, and action, an effective model should neither under-reason complex questions nor over-reason simple ones. To evaluate this property, we measure the alignment between the generated reasoning depth and the expected depth implied by each question. As shown in Figure~\ref{plus-fig:reasoning-depth}, most baseline models exhibit relatively low thinking depth accuracy, suggesting that their reasoning behaviors are largely fixed, often producing incomplete reasoning for complex questions or unnecessarily verbose reasoning for simpler ones, regardless of the semantic demands of the task. Such mismatches indicate limited adaptability and may negatively affect both reasoning quality and response efficiency in practical scenarios. In contrast, Vad-R1-Plus achieves the highest reasoning depth score among all compared models. This result demonstrates that Vad-R1-Plus can reliably adjust its reasoning depth in response to different question types, generating concise reasoning for perception-oriented queries while producing more elaborate, multi-stage analysis for cognition- and action-oriented questions.

\begin{figure*}[t]
  \centering
  \begin{subfigure}{0.24\textwidth}
    \centering
    \includegraphics[width=\linewidth]{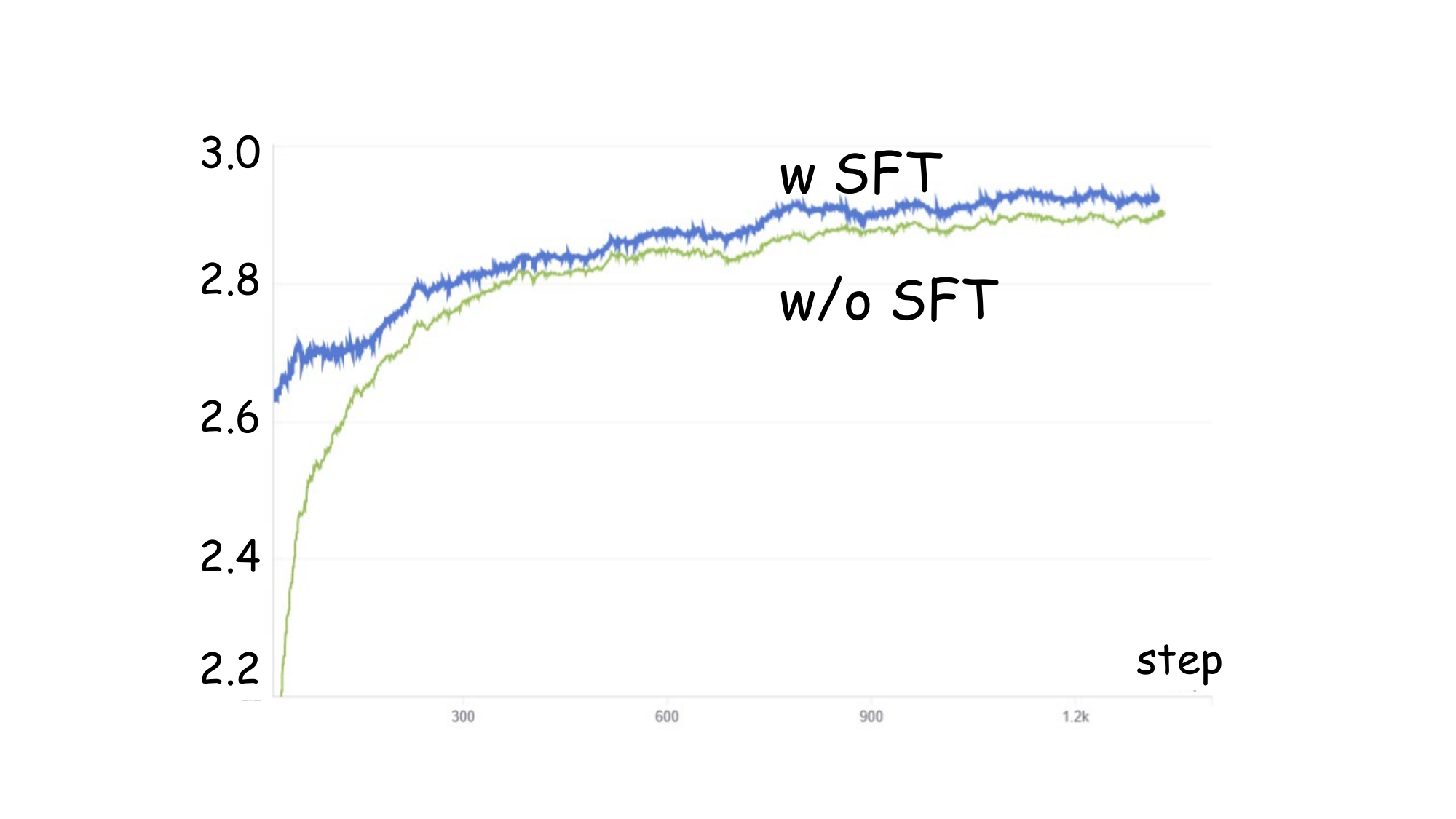}
    \caption{Total reward.}
  \end{subfigure}
  \hfill
  \begin{subfigure}{0.24\textwidth}
    \centering
    \includegraphics[width=\linewidth]{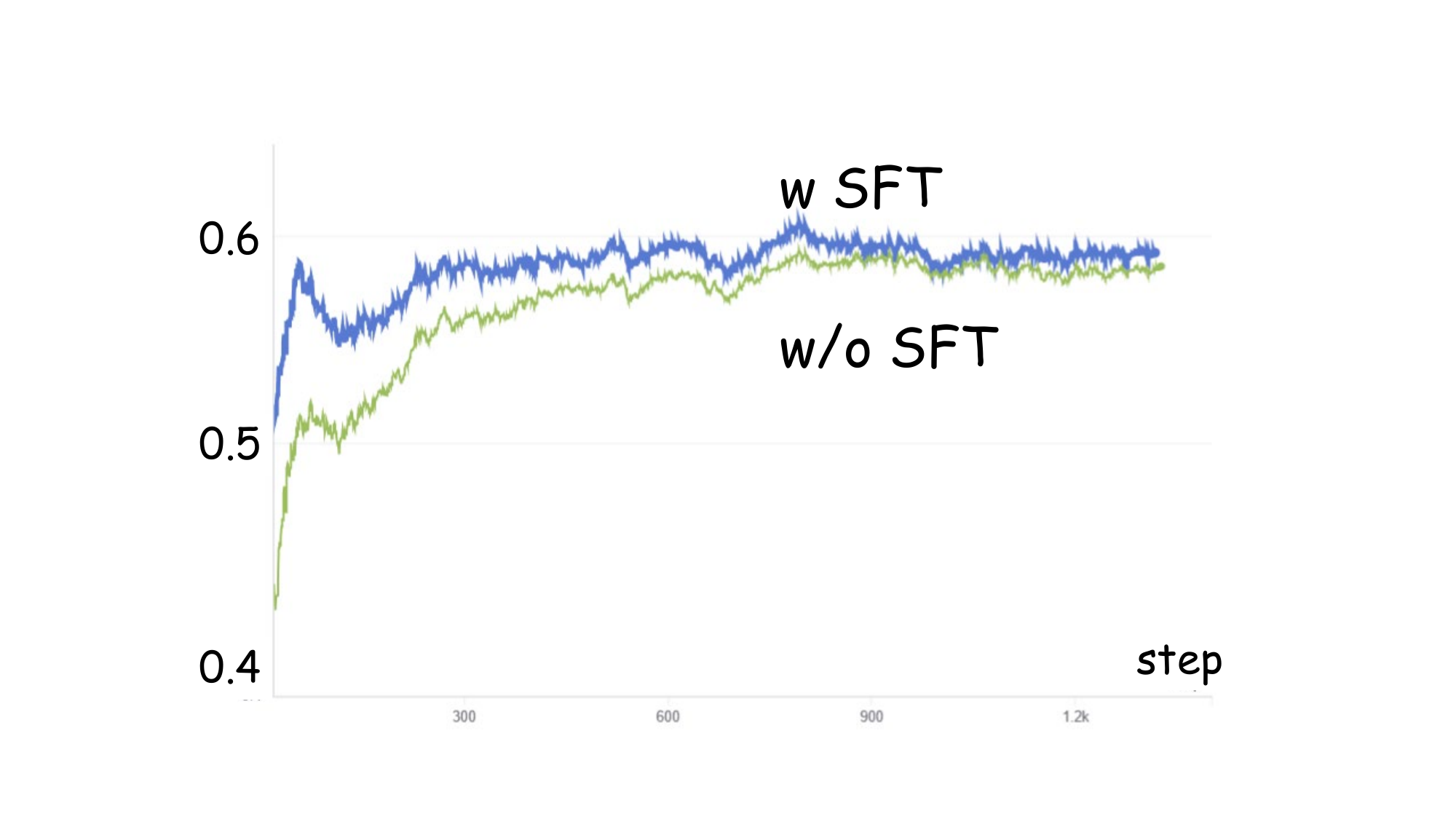}
    \caption{Accuracy reward.}
  \end{subfigure}
  \hfill
  \begin{subfigure}{0.24\textwidth}
    \centering
    \includegraphics[width=\linewidth]{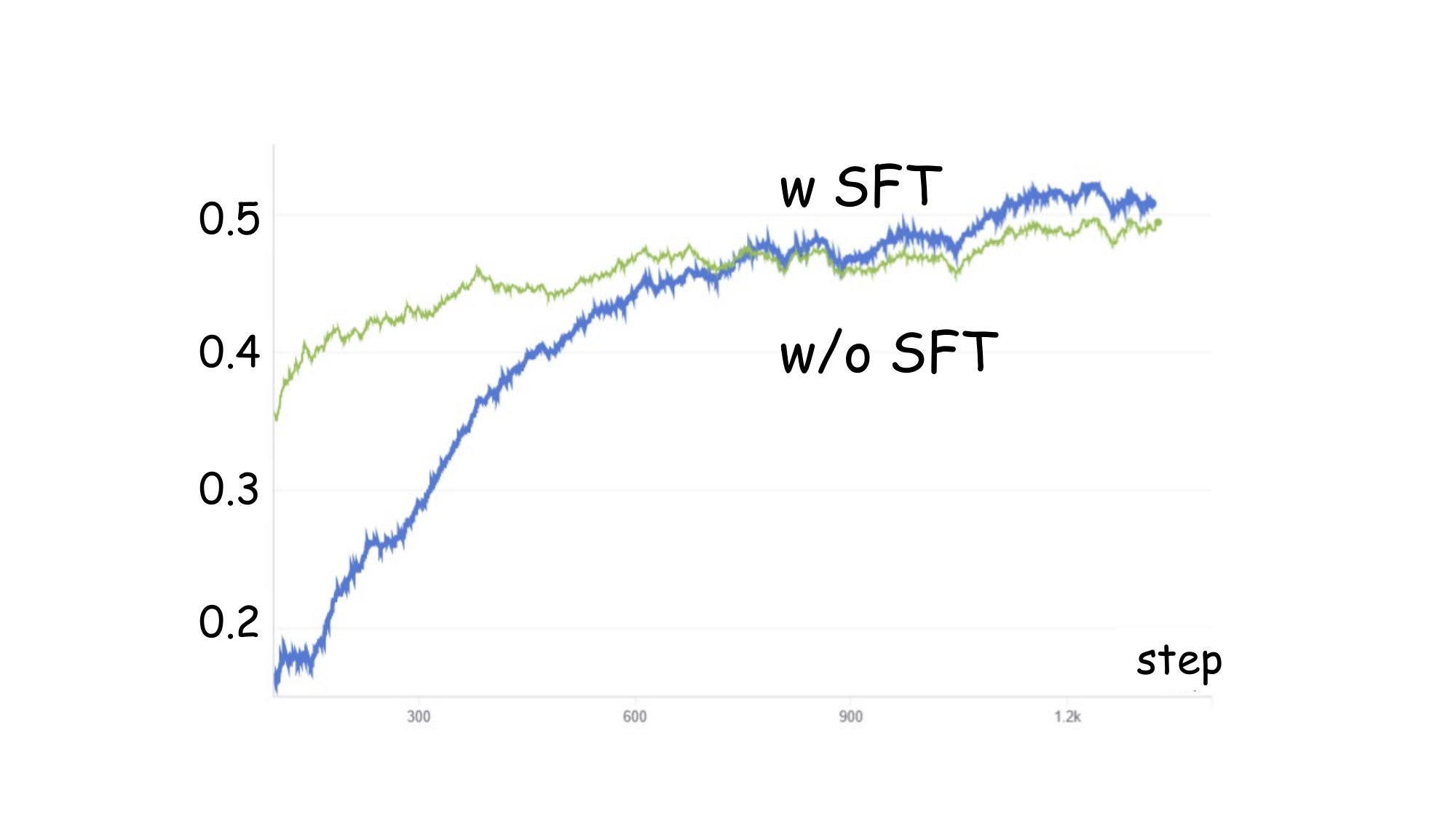}
    \caption{Risk reward.}
  \end{subfigure}
  \hfill
  \begin{subfigure}{0.24\textwidth}
    \centering
    \includegraphics[width=\linewidth]{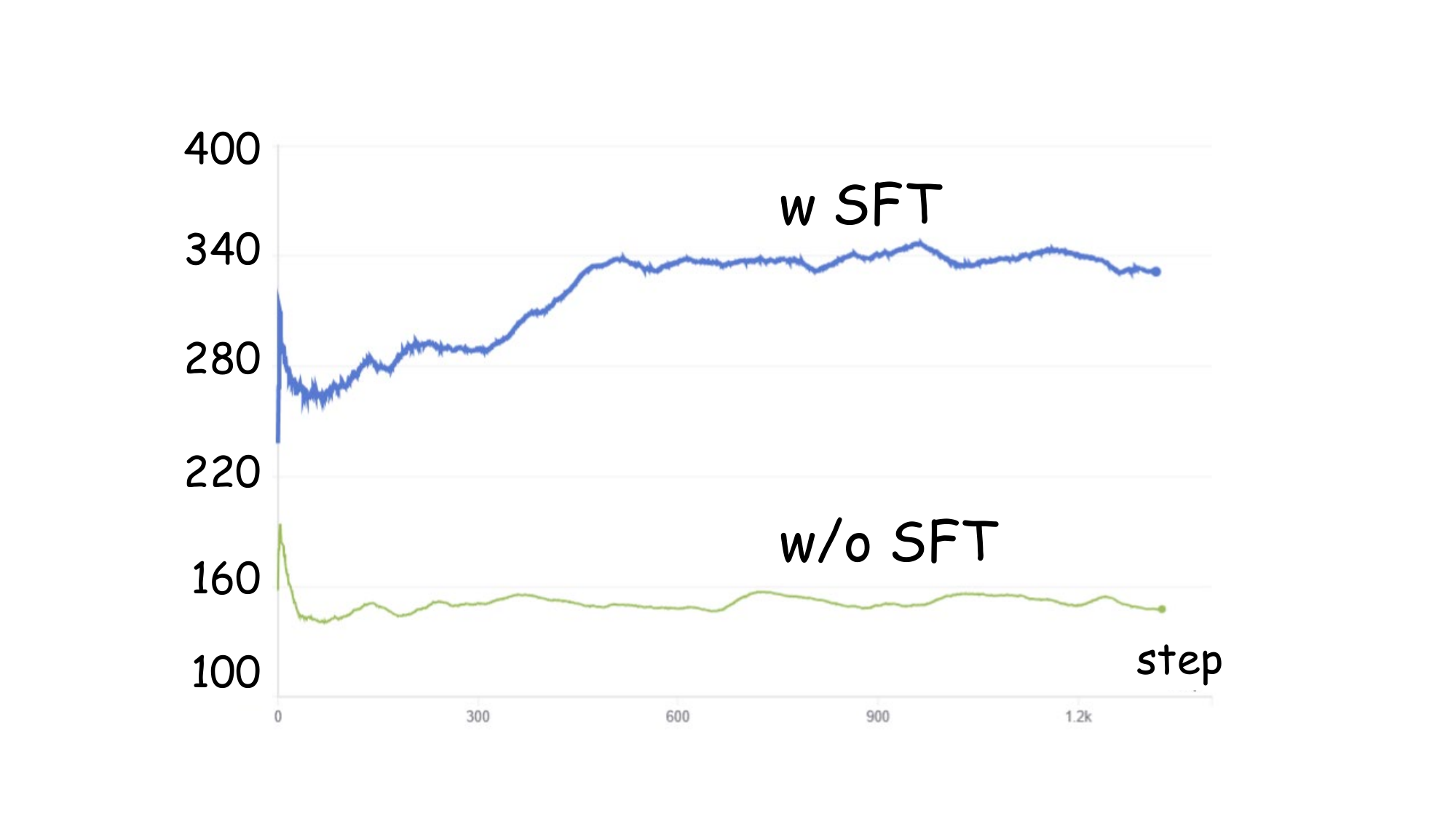}
    \caption{Output length.}
  \end{subfigure}
  \caption{RL training curves of Vad-R1-Plus. The blue curves represent RL initialized from the SFT-trained model, while the green curves denote RL applied directly to the base model without SFT. }
  \label{plus-fig:training-curve}
\end{figure*}

\subsubsection{Reliability Evaluation}

To further assess the reliability of video anomaly reasoning, we introduce \textbf{Double Right} and \textbf{Triple Right} metrics, which jointly evaluate the correctness of reasoning and final answers, requiring explicit consistency between the intermediate reasoning process and the final decision. Specifically, Double Right measures the proportion of cases where both the reasoning process and the final answer are simultaneously correct. This metric reflects whether a model arrives at correct answers through logically sound reasoning, rather than by coincidence or superficial pattern matching.

As shown in the left part of Table~\ref{plus-tab:double-right}, Vad-R1-Plus achieves the highest Double Right score among all compared methods, indicating a stronger ability to produce answers that are reliably supported by valid reasoning. We then consider the more stringent Triple Right metric, which additionally requires the anomaly category predicted during the reasoning process to be correct. Under this stricter criterion, the performance gap between Vad-R1-Plus and existing baselines becomes substantially larger. Although several strong baselines maintain relatively competitive Double Right scores, their Triple Right scores drop sharply. This behavior suggests that, while these models may occasionally produce correct answers with plausible reasoning, their reasoning processes are often not grounded in correct anomaly semantics. In contrast, Vad-R1-Plus consistently maintains high performance under Triple Right, demonstrating its ability to correctly integrate visual evidence, anomaly interpretation, category prediction, and final decision.

\subsubsection{Training Curves}

We further visualize the RL dynamics of Vad-R1-Plus in Figure~\ref{plus-fig:training-curve} by comparing models trained with and without supervised fine-tuning (SFT) initialization. Overall, models initialized with SFT exhibit more stable and consistently superior training behavior across all metrics. As shown in Figure~\ref{plus-fig:training-curve}(a) and (b), the SFT-initialized model achieves higher total reward and accuracy reward throughout training, indicating that supervised reasoning priors provide a strong foundation for effective policy optimization. In contrast, the model trained without SFT struggles to improve accuracy and converges to a significantly lower reward level. As shown in Figure~\ref{plus-fig:training-curve}(c), the model trained without SFT attains a slightly higher risk reward at the very early stage of RL. This behavior is primarily due to the coarse-grained nature of risk supervision, which can be temporarily satisfied by shortcut strategies such as biased risk guessing without grounded reasoning. In contrast, the SFT-initialized model follows a more conservative but stable optimization trajectory at the beginning, as it is constrained by structured reasoning patterns learned during SFT. As training progresses, it consistently surpasses the model without SFT, demonstrating that risk assessment grounded in coherent anomaly reasoning leads to more reliable optimization. Furthermore, the output length statistics in Figure~\ref{plus-fig:training-curve}(d) reveal that SFT provides a strong structural prior for reasoning generation. Models trained with SFT maintain longer and more stable outputs during RL, whereas models trained without SFT tend to collapse to shorter responses, reflecting insufficient reasoning depth. Overall, these results highlight the critical role of supervised fine-tuning in stabilizing reinforcement learning and enabling effective optimization of anomaly-aware rewards.

\subsubsection{Qualitative Analyses}

\begin{figure*}[t]
    \centering
    \includegraphics[width=1\linewidth]{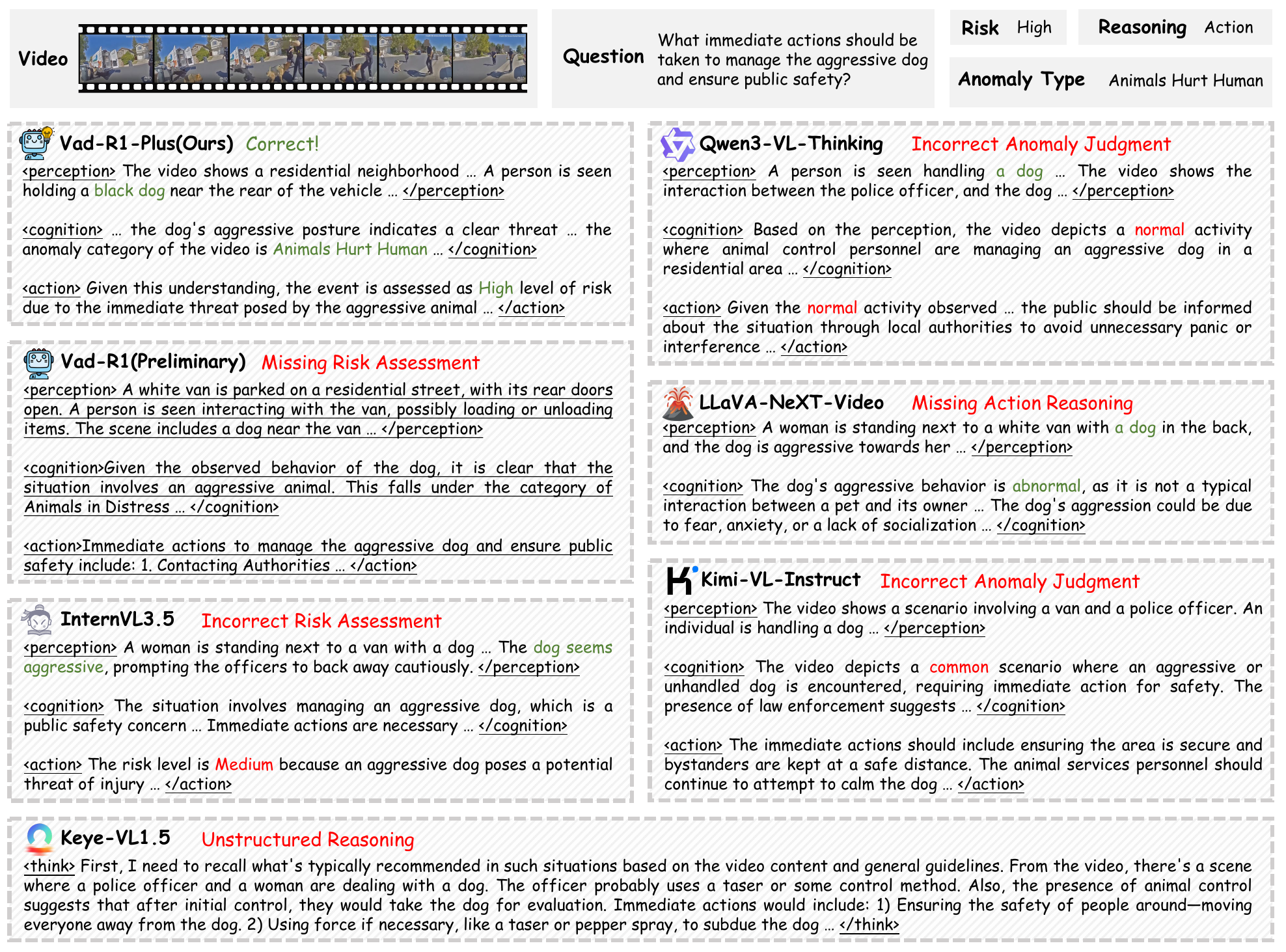} 
    \caption{Qualitative comparison.}
    \label{plus-fig:visualization}
\end{figure*}

Figure~\ref{plus-fig:visualization} presents a qualitative comparison on a case involving an aggressive dog that poses a high public risk. Vad-R1-Plus produces a complete reasoning chain, correctly identifies the aggressive behavior in perception stage, accurately classifies the anomaly as \emph{Animals Hurt Human} in cognition stage, and assigns a high risk level with appropriate emergency responses in action stage, demonstrating coherent cross-stage reasoning and strong alignment among intermediate analysis. In contrast, the preliminary version Vad-R1 fails to explicitly assess the risk level of the event. This indicates that without explicit risk modeling, anomaly reasoning remains insufficient for safety-critical scenarios, even when the anomaly is correctly identified. Other baseline models fail in different ways. InternVL3.5~\cite{wang2025internvl3} correctly perceives the aggressive behavior but underestimates its severity, leading to an incorrect medium-risk assessment. Qwen3-VL-Thinking~\cite{Qwen3vl} misjudges the event as normal, which undermines subsequent reasoning. LLaVA-NeXT-Video~\cite{zhang2024video} identifies the anomaly but fails to generate the action stage, leaving the safety decision incomplete. Kimi-VL-Instruct~\cite{team2025kimi} further exhibits incorrect anomaly judgment, while Keye-VL1.5~\cite{yang2025kwai} produces unstructured reasoning without clear stage separation. Overall, this comparison highlights that effective video anomaly reasoning requires not only recognizing abnormal events, but also explicitly assessing risk and generating actionable decisions.

\section{Conclusion}

In this paper, we re-examine video anomaly understanding from a reasoning-oriented perspective and argue that detection-centric formulations are insufficient for safety-critical scenarios. To address this limitation, we define Video Anomaly Reasoning (VAR) as a new task that requires models to reason about abnormal events in terms of causes, risks, and appropriate responses. To support systematic research on VAR, we present Vad-Reasoning-Plus, a large-scale benchmark with structured, stage-aware annotations spanning perception, cognition, and action. Combined with the proposed Perception–Cognition–Action Chain-of-Thought and the A\textsuperscript{2}-GRPO training strategy, this benchmark enables principled evaluation and learning of structured, adaptive, and risk-aware reasoning behaviors. Extensive experiments demonstrate that models trained under this framework consistently outperform strong open-source and proprietary baselines in both reasoning quality and answer reliability. We believe that this work establishes a foundation for reasoning-centric video anomaly understanding, encouraging future research to move beyond detection-focused paradigms toward more explainable, risk-aware, and decision-oriented video intelligence systems that can be effectively integrated into real-world applications.


\clearpage

{
\bibliographystyle{IEEEtran}
\bibliography{references}

@article{tang2024hawk,
  title={Hawk: Learning to understand open-world video anomalies},
  author={Tang, Jiaqi and Lu, Hao and Wu, Ruizheng and Xu, Xiaogang and Ma, Ke and Fang, Cheng and Guo, Bin and Lu, Jiangbo and Chen, Qifeng and Chen, Yingcong},
  journal={Advances in Neural Information Processing Systems},
  volume={37},
  pages={139751--139785},
  year={2024}
}

@article{lv2021localizing,
  title={Localizing anomalies from weakly-labeled videos},
  author={Lv, Hui and Zhou, Chuanwei and Cui, Zhen and Xu, Chunyan and Li, Yong and Yang, Jian},
  journal={IEEE transactions on image processing},
  volume={30},
  pages={4505--4515},
  year={2021},
  publisher={IEEE}
}

@inproceedings{wu2020not,
  title={Not only look, but also listen: Learning multimodal violence detection under weak supervision},
  author={Wu, Peng and Liu, Jing and Shi, Yujia and Sun, Yujia and Shao, Fangtao and Wu, Zhaoyang and Yang, Zhiwei},
  booktitle={Computer Vision--ECCV 2020: 16th European Conference, Glasgow, UK, August 23--28, 2020, Proceedings, Part XXX 16},
  pages={322--339},
  year={2020},
  organization={Springer}
}

@inproceedings{sultani2018real,
  title={Real-world anomaly detection in surveillance videos},
  author={Sultani, Waqas and Chen, Chen and Shah, Mubarak},
  booktitle={Proceedings of the IEEE conference on computer vision and pattern recognition},
  pages={6479--6488},
  year={2018}
}

@inproceedings{wang2025federated,
  title={Federated Weakly Supervised Video Anomaly Detection with Multimodal Prompt},
  author={Wang, Benfeng and Huang, Chao and Wen, Jie and Wang, Wei and Liu, Yabo and Xu, Yong},
  booktitle={Proceedings of the AAAI Conference on Artificial Intelligence},
  volume={39},
  number={20},
  pages={21017--21025},
  year={2025}
}

@article{luo2021future,
  title={Future frame prediction network for video anomaly detection},
  author={Luo, Weixin and Liu, Wen and Lian, Dongze and Gao, Shenghua},
  journal={IEEE transactions on pattern analysis and machine intelligence},
  volume={44},
  number={11},
  pages={7505--7520},
  year={2021},
  publisher={IEEE}
}

@inproceedings{huang2024long,
  title={Long short-term dynamic prototype alignment learning for video anomaly detection},
  author={Huang, Chao and Wen, Jie and Liu, Chengliang and Liu, Yabo},
  booktitle={Proceedings of the Thirty-Third International Joint Conference on Artificial Intelligence},
  pages={866--874},
  year={2024}
}

@article{huang2022weakly,
  title={Weakly supervised video anomaly detection via self-guided temporal discriminative transformer},
  author={Huang, Chao and Liu, Chengliang and Wen, Jie and Wu, Lian and Xu, Yong and Jiang, Qiuping and Wang, Yaowei},
  journal={IEEE Transactions on Cybernetics},
  volume={54},
  number={5},
  pages={3197--3210},
  year={2022},
  publisher={IEEE}
}

@article{huang2022self,
  title={Self-supervised attentive generative adversarial networks for video anomaly detection},
  author={Huang, Chao and Wen, Jie and Xu, Yong and Jiang, Qiuping and Yang, Jian and Wang, Yaowei and Zhang, David},
  journal={IEEE transactions on neural networks and learning systems},
  volume={34},
  number={11},
  pages={9389--9403},
  year={2022},
  publisher={IEEE}
}

@article{huang2021abnormal,
  title={Abnormal event detection using deep contrastive learning for intelligent video surveillance system},
  author={Huang, Chao and Wu, Zhihao and Wen, Jie and Xu, Yong and Jiang, Qiuping and Wang, Yaowei},
  journal={IEEE Transactions on Industrial Informatics},
  volume={18},
  number={8},
  pages={5171--5179},
  year={2021},
  publisher={IEEE}
}

@article{huang2025multimodal,
  title={Multimodal Evidential Learning for Open-World Weakly-Supervised Video Anomaly Detection},
  author={Huang, Chao and Huang, Weiliang and Jiang, Qiuping and Wang, Wei and Wen, Jie and Zhang, Bob},
  journal={IEEE Transactions on Multimedia},
  year={2025},
  publisher={IEEE}
}

@inproceedings{wu2024vadclip,
  title={Vadclip: Adapting vision-language models for weakly supervised video anomaly detection},
  author={Wu, Peng and Zhou, Xuerong and Pang, Guansong and Zhou, Lingru and Yan, Qingsen and Wang, Peng and Zhang, Yanning},
  booktitle={Proceedings of the AAAI Conference on Artificial Intelligence},
  volume={38},
  number={6},
  pages={6074--6082},
  year={2024}
}

@inproceedings{ye2025vera,
  title={Vera: Explainable video anomaly detection via verbalized learning of vision-language models},
  author={Ye, Muchao and Liu, Weiyang and He, Pan},
  booktitle={Proceedings of the Computer Vision and Pattern Recognition Conference},
  pages={8679--8688},
  year={2025}
}

@article{lin2023video,
  title={Video-llava: Learning united visual representation by alignment before projection},
  author={Lin, Bin and Ye, Yang and Zhu, Bin and Cui, Jiaxi and Ning, Munan and Jin, Peng and Yuan, Li},
  journal={arXiv preprint arXiv:2311.10122},
  year={2023}
}

@article{li2023videochat,
  title={Videochat: Chat-centric video understanding},
  author={Li, KunChang and He, Yinan and Wang, Yi and Li, Yizhuo and Wang, Wenhai and Luo, Ping and Wang, Yali and Wang, Limin and Qiao, Yu},
  journal={arXiv preprint arXiv:2305.06355},
  year={2023}
}

@article{lv2024video,
  title={Video anomaly detection and explanation via large language models},
  author={Lv, Hui and Sun, Qianru},
  journal={arXiv preprint arXiv:2401.05702},
  year={2024}
}

@article{zhang2024holmes,
  title={Holmes-vad: Towards unbiased and explainable video anomaly detection via multi-modal llm},
  author={Zhang, Huaxin and Xu, Xiaohao and Wang, Xiang and Zuo, Jialong and Han, Chuchu and Huang, Xiaonan and Gao, Changxin and Wang, Yuehuan and Sang, Nong},
  journal={arXiv preprint arXiv:2406.12235},
  year={2024}
}

@article{zhang2024holmesvau,
  title={Holmes-vau: Towards long-term video anomaly understanding at any granularity},
  author={Zhang, Huaxin and Xu, Xiaohao and Wang, Xiang and Zuo, Jialong and Huang, Xiaonan and Gao, Changxin and Zhang, Shanjun and Yu, Li and Sang, Nong},
  journal={arXiv preprint arXiv:2412.06171},
  year={2024}
}

@inproceedings{du2024uncovering,
  title={Uncovering what why and how: A comprehensive benchmark for causation understanding of video anomaly},
  author={Du, Hang and Zhang, Sicheng and Xie, Binzhu and Nan, Guoshun and Zhang, Jiayang and Xu, Junrui and Liu, Hangyu and Leng, Sicong and Liu, Jiangming and Fan, Hehe and others},
  booktitle={Proceedings of the IEEE/CVF Conference on Computer Vision and Pattern Recognition},
  pages={18793--18803},
  year={2024}
}

@article{du2024exploring,
  title={Exploring What Why and How: A Multifaceted Benchmark for Causation Understanding of Video Anomaly},
  author={Du, Hang and Nan, Guoshun and Qian, Jiawen and Wu, Wangchenhui and Deng, Wendi and Mu, Hanqing and Chen, Zhenyan and Mao, Pengxuan and Tao, Xiaofeng and Liu, Jun},
  journal={arXiv preprint arXiv:2412.07183},
  year={2024}
}

@inproceedings{ma2025sherlock,
  title={Sherlock: Towards Multi-scene Video Abnormal Event Extraction and Localization via a Global-local Spatial-sensitive LLM},
  author={Ma, Junxiao and Wang, Jingjing and Luo, Jiamin and Yu, Peiying and Zhou, Guodong},
  booktitle={Proceedings of the ACM on Web Conference 2025},
  pages={4004--4013},
  year={2025}
}

@article{jaech2024openai,
  title={Openai o1 system card},
  author={{OpenAI}},
  journal={arXiv preprint arXiv:2412.16720},
  year={2024}
}

@article{guo2025deepseek,
  title={{DeepSeek-R1}: Incentivizing reasoning capability in llms via reinforcement learning},
  author={{DeepSeek-AI}},
  journal={arXiv preprint arXiv:2501.12948},
  year={2025}
}

@misc{QwenQvQ,
  title={{QVQ-Max}: Think with Evidence},
  author={{Qwen Team}},
  year={2025},
  url={https://qwenlm.github.io/blog/qvq-max-preview/}
}

@article{yang2025qwen3,
  title={{Qwen3 technical report}},
  author={{Qwen Team}},
  journal={arXiv preprint arXiv:2505.09388},
  year={2025}
}

@article{shao2024deepseekmath,
  title={Deepseekmath: Pushing the limits of mathematical reasoning in open language models},
  author={Shao, Zhihong and Wang, Peiyi and Zhu, Qihao and Xu, Runxin and Song, Junxiao and Bi, Xiao and Zhang, Haowei and Zhang, Mingchuan and Li, YK and Wu, Y and others},
  journal={arXiv preprint arXiv:2402.03300},
  year={2024}
}

@article{yu2025unhackable,
  title={Unhackable temporal rewarding for scalable video mllms},
  author={Yu, En and Lin, Kangheng and Zhao, Liang and Wei, Yana and Zhu, Zining and Wei, Haoran and Sun, Jianjian and Ge, Zheng and Zhang, Xiangyu and Wang, Jingyu and others},
  journal={arXiv preprint arXiv:2502.12081},
  year={2025}
}

@inproceedings{yang2024follow,
  title={Follow the rules: reasoning for video anomaly detection with large language models},
  author={Yang, Yuchen and Lee, Kwonjoon and Dariush, Behzad and Cao, Yinzhi and Lo, Shao-Yuan},
  booktitle={European Conference on Computer Vision},
  pages={304--322},
  year={2024},
  organization={Springer}
}

@inproceedings{zanella2024harnessing,
  title={Harnessing large language models for training-free video anomaly detection},
  author={Zanella, Luca and Menapace, Willi and Mancini, Massimiliano and Wang, Yiming and Ricci, Elisa},
  booktitle={Proceedings of the IEEE/CVF Conference on Computer Vision and Pattern Recognition},
  pages={18527--18536},
  year={2024}
}

@inproceedings{chen2023tevad,
  title={TEVAD: Improved video anomaly detection with captions},
  author={Chen, Weiling and Ma, Keng Teck and Yew, Zi Jian and Hur, Minhoe and Khoo, David Aik-Aun},
  booktitle={Proceedings of the IEEE/CVF Conference on Computer Vision and Pattern Recognition},
  pages={5549--5559},
  year={2023}
}

@inproceedings{acsintoae2022ubnormal,
  title={Ubnormal: New benchmark for supervised open-set video anomaly detection},
  author={Acsintoae, Andra and Florescu, Andrei and Georgescu, Mariana-Iuliana and Mare, Tudor and Sumedrea, Paul and Ionescu, Radu Tudor and Khan, Fahad Shahbaz and Shah, Mubarak},
  booktitle={Proceedings of the IEEE/CVF conference on computer vision and pattern recognition},
  pages={20143--20153},
  year={2022}
}

@inproceedings{yuan2024towards,
  title={Towards surveillance video-and-language understanding: New dataset baselines and challenges},
  author={Yuan, Tongtong and Zhang, Xuange and Liu, Kun and Liu, Bo and Chen, Chen and Jin, Jian and Jiao, Zhenzhen},
  booktitle={Proceedings of the IEEE/CVF Conference on Computer Vision and Pattern Recognition},
  pages={22052--22061},
  year={2024}
}

@inproceedings{li2024llama,
  title={Llama-vid: An image is worth 2 tokens in large language models},
  author={Li, Yanwei and Wang, Chengyao and Jia, Jiaya},
  booktitle={European Conference on Computer Vision},
  pages={323--340},
  year={2024},
  organization={Springer}
}

@article{zhang2024long,
  title={Long context transfer from language to vision},
  author={Zhang, Peiyuan and Zhang, Kaichen and Li, Bo and Zeng, Guangtao and Yang, Jingkang and Zhang, Yuanhan and Wang, Ziyue and Tan, Haoran and Li, Chunyuan and Liu, Ziwei},
  journal={arXiv preprint arXiv:2406.16852},
  year={2024}
}

@inproceedings{jin2024chat,
  title={Chat-univi: Unified visual representation empowers large language models with image and video understanding},
  author={Jin, Peng and Takanobu, Ryuichi and Zhang, Wancai and Cao, Xiaochun and Yuan, Li},
  booktitle={Proceedings of the IEEE/CVF Conference on Computer Vision and Pattern Recognition},
  pages={13700--13710},
  year={2024}
}

@article{di2025streaming,
  title={Streaming video question-answering with in-context video kv-cache retrieval},
  author={Di, Shangzhe and Yu, Zhelun and Zhang, Guanghao and Li, Haoyuan and Zhong, Tao and Cheng, Hao and Li, Bolin and He, Wanggui and Shu, Fangxun and Jiang, Hao},
  journal={arXiv preprint arXiv:2503.00540},
  year={2025}
}

@article{xiong2025streaming,
  title={Streaming Video Understanding and Multi-round Interaction with Memory-enhanced Knowledge},
  author={Xiong, Haomiao and Yang, Zongxin and Yu, Jiazuo and Zhuge, Yunzhi and Zhang, Lu and Zhu, Jiawen and Lu, Huchuan},
  journal={arXiv preprint arXiv:2501.13468},
  year={2025}
}

@article{xu2024llava,
  title={Llava-o1: Let vision language models reason step-by-step},
  author={Xu, Guowei and Jin, Peng and Hao, Li and Song, Yibing and Sun, Lichao and Yuan, Li},
  journal={arXiv preprint arXiv:2411.10440},
  year={2024}
}

@article{thawakar2025llamav,
  title={Llamav-o1: Rethinking step-by-step visual reasoning in llms},
  author={Thawakar, Omkar and Dissanayake, Dinura and More, Ketan and Thawkar, Ritesh and Heakl, Ahmed and Ahsan, Noor and Li, Yuhao and Zumri, Mohammed and Lahoud, Jean and Anwer, Rao Muhammad and others},
  journal={arXiv preprint arXiv:2501.06186},
  year={2025}
}

@article{liu2025videomind,
  title={VideoMind: A Chain-of-LoRA Agent for Long Video Reasoning},
  author={Liu, Ye and Lin, Kevin Qinghong and Chen, Chang Wen and Shou, Mike Zheng},
  journal={arXiv preprint arXiv:2503.13444},
  year={2025}
}

@article{huang2025vision,
  title={Vision-r1: Incentivizing reasoning capability in multimodal large language models},
  author={Huang, Wenxuan and Jia, Bohan and Zhai, Zijie and Cao, Shaosheng and Ye, Zheyu and Zhao, Fei and Xu, Zhe and Hu, Yao and Lin, Shaohui},
  journal={arXiv preprint arXiv:2503.06749},
  year={2025}
}

@article{wang2025timezero,
  title={TimeZero: Temporal Video Grounding with Reasoning-Guided LVLM},
  author={Wang, Ye and Xu, Boshen and Yue, Zihao and Xiao, Zihan and Wang, Ziheng and Zhang, Liang and Yang, Dingyi and Wang, Wenxuan and Jin, Qin},
  journal={arXiv preprint arXiv:2503.13377},
  year={2025}
}

@article{li2025videochat,
  title={VideoChat-R1: Enhancing Spatio-Temporal Perception via Reinforcement Fine-Tuning},
  author={Li, Xinhao and Yan, Ziang and Meng, Desen and Dong, Lu and Zeng, Xiangyu and He, Yinan and Wang, Yali and Qiao, Yu and Wang, Yi and Wang, Limin},
  journal={arXiv preprint arXiv:2504.06958},
  year={2025}
}

@article{feng2025video,
  title={Video-r1: Reinforcing video reasoning in mllms},
  author={Feng, Kaituo and Gong, Kaixiong and Li, Bohao and Guo, Zonghao and Wang, Yibing and Peng, Tianshuo and Wang, Benyou and Yue, Xiangyu},
  journal={arXiv preprint arXiv:2503.21776},
  year={2025}
}

@article{bai2025qwen25vl,
  title={Qwen2.5-vl technical report},
  author={{Qwen Team}},
  journal={arXiv preprint arXiv:2502.13923},
  year={2025}
}

@article{Qwen3vl,
      title={Qwen3-VL Technical Report}, 
      author={{Qwen Team}},
	  journal={arXiv preprint arXiv:2511.21631},
      year={2025}
}

@article{team2025kimi,
  title={Kimi-vl technical report},
  author={{Kimi Team}},
  journal={arXiv preprint arXiv:2504.07491},
  year={2025}
}

@inproceedings{shu2025video,
  title={Video-xl: Extra-long vision language model for hour-scale video understanding},
  author={Shu, Yan and Liu, Zheng and Zhang, Peitian and Qin, Minghao and Zhou, Junjie and Liang, Zhengyang and Huang, Tiejun and Zhao, Bo},
  booktitle={Proceedings of the Computer Vision and Pattern Recognition Conference},
  pages={26160--26169},
  year={2025}
}

@inproceedings{papineni2002bleu,
  title={Bleu: a method for automatic evaluation of machine translation},
  author={Papineni, Kishore and Roukos, Salim and Ward, Todd and Zhu, Wei-Jing},
  booktitle={Proceedings of the 40th annual meeting of the Association for Computational Linguistics},
  pages={311--318},
  year={2002}
}

@inproceedings{banerjee2005meteor,
  title={METEOR: An automatic metric for MT evaluation with improved correlation with human judgments},
  author={Banerjee, Satanjeev and Lavie, Alon},
  booktitle={Proceedings of the acl workshop on intrinsic and extrinsic evaluation measures for machine translation and/or summarization},
  pages={65--72},
  year={2005}
}

@inproceedings{lin2004rouge,
  title={Rouge: A package for automatic evaluation of summaries},
  author={Lin, Chin-Yew},
  booktitle={Text summarization branches out},
  pages={74--81},
  year={2004}
}

@inproceedings{gani2025vane,
  title={VANE-Bench: Video Anomaly Evaluation Benchmark for Conversational LMMs},
  author={Gani, Hanan and Bharadwaj, Rohit and Naseer, Muzammal and Khan, Fahad Shahbaz and Khan, Salman},
  booktitle={Findings of the Association for Computational Linguistics: NAACL 2025},
  pages={3123--3140},
  year={2025}
}

@article{wang2025internvl3,
  title={Internvl3.5: Advancing open-source multimodal models in versatility, reasoning, and efficiency},
  author={Wang, Weiyun and Gao, Zhangwei and Gu, Lixin and Pu, Hengjun and Cui, Long and Wei, Xingguang and Liu, Zhaoyang and Jing, Linglin and Ye, Shenglong and Shao, Jie and others},
  journal={arXiv preprint arXiv:2508.18265},
  year={2025}
}

@article{wang2025internvideo2,
  title={InternVideo2.5: Empowering Video MLLMs with Long and Rich Context Modeling},
  author={Wang, Yi and Li, Xinhao and Yan, Ziang and He, Yinan and Yu, Jiashuo and Zeng, Xiangyu and Wang, Chenting and Ma, Changlian and Huang, Haian and Gao, Jianfei and others},
  journal={arXiv preprint arXiv:2501.12386},
  year={2025}
}

@article{li2024videochat,
  title={Videochat-flash: Hierarchical compression for long-context video modeling},
  author={Li, Xinhao and Wang, Yi and Yu, Jiashuo and Zeng, Xiangyu and Zhu, Yuhan and Huang, Haian and Gao, Jianfei and Li, Kunchang and He, Yinan and Wang, Chenting and others},
  journal={arXiv preprint arXiv:2501.00574},
  year={2024}
}

@article{zhang2025videollama,
  title={VideoLLaMA 3: Frontier Multimodal Foundation Models for Image and Video Understanding},
  author={Zhang, Boqiang and Li, Kehan and Cheng, Zesen and Hu, Zhiqiang and Yuan, Yuqian and Chen, Guanzheng and Leng, Sicong and Jiang, Yuming and Zhang, Hang and Li, Xin and others},
  journal={arXiv preprint arXiv:2501.13106},
  year={2025}
}

@article{zhang2024video,
  title={Video instruction tuning with synthetic data},
  author={Zhang, Yuanhan and Wu, Jinming and Li, Wei and Li, Bo and Ma, Zejun and Liu, Ziwei and Li, Chunyuan},
  journal={arXiv preprint arXiv:2410.02713},
  year={2024}
}

@article{li2024llavaone,
  title={Llava-onevision: Easy visual task transfer},
  author={Li, Bo and Zhang, Yuanhan and Guo, Dong and Zhang, Renrui and Li, Feng and Zhang, Hao and Zhang, Kaichen and Zhang, Peiyuan and Li, Yanwei and Liu, Ziwei and others},
  journal={arXiv preprint arXiv:2408.03326},
  year={2024}
}

@misc{wang-2025-open-r1-video,
  author = {Xiaodong Wang and Peixi Peng},
  title = {Open-R1-Video},
  year = {2025},
  url = {https://github.com/Wang-Xiaodong1899/Open-R1-Video}
}

@article{yang2025kwai,
  title={Kwai keye-vl 1.5 technical report},
  author={{keye Team}},
  journal={arXiv preprint arXiv:2509.01563},
  year={2025}
}

@article{xiaomi2025mimo,
  title={Mimo-vl technical report},
  author={{LLM-Core Xiaomi}},
  journal={arXiv preprint arXiv:2506.03569},
  year={2025}
}

@article{yu2025minicpm,
  title={Minicpm-v 4.5: Cooking efficient mllms via architecture, data, and training recipe},
  author={{MiniCPM-V Team}},
  journal={arXiv preprint arXiv:2509.18154},
  year={2025}
}

@article{lu2025ovis2,
  title={Ovis2. 5 technical report},
  author={{Ovis Team}},
  journal={arXiv preprint arXiv:2508.11737},
  year={2025}
}

@article{v2507glm,
  title={Glm-4.5 v and glm-4.1 v-thinking: Towards versatile multimodal reasoning with scalable reinforcement learning, 2025},
  author={{GLM-V Team}},
  journal={arXiv preprint arXiv:2507.01006},
  year={2025}
}

@article{comanici2025gemini,
  title={Gemini 2.5: Pushing the frontier with advanced reasoning, multimodality, long context, and next generation agentic capabilities},
  author={{Gemini Team}},
  journal={arXiv preprint arXiv:2507.06261},
  year={2025}
}

@misc{grok4,
  title        = {Grok 4 Model},
  author       = {{xAI}},
  year         = {2025},
  url = {https://x.ai/news/grok-4}
}

@misc{gpt5,
  title        = {GPT-5 System Card},
  author       = {{OpenAI}},
  year         = {2025},
  url = {https://cdn.openai.com/gpt-5-system-card.pdf},
}

@inproceedings{lu2013abnormal,
  title={Abnormal event detection at 150 fps in matlab},
  author={Lu, Cewu and Shi, Jianping and Jia, Jiaya},
  booktitle={Proceedings of the IEEE international conference on computer vision},
  pages={2720--2727},
  year={2013}
}

@article{li2013anomaly,
  title={Anomaly detection and localization in crowded scenes},
  author={Li, Weixin and Mahadevan, Vijay and Vasconcelos, Nuno},
  journal={IEEE transactions on pattern analysis and machine intelligence},
  volume={36},
  number={1},
  pages={18--32},
  year={2013},
  publisher={IEEE}
}

@article{cao2024scene,
  title={Scene-dependent prediction in latent space for video anomaly detection and anticipation},
  author={Cao, Congqi and Zhang, Hanwen and Lu, Yue and Wang, Peng and Zhang, Yanning},
  journal={IEEE transactions on pattern analysis and machine intelligence},
  year={2024},
  publisher={IEEE}
}

@inproceedings{zhao2025smarthome,
  title={SmartHome-Bench: A Comprehensive Benchmark for Video Anomaly Detection in Smart Homes Using Multi-Modal Large Language Models},
  author={Zhao, Xinyi and Zhang, Congjing and Guo, Pei and Li, Wei and Chen, Lin and Zhao, Chaoyue and Huang, Shuai},
  booktitle={Proceedings of the Computer Vision and Pattern Recognition Conference},
  pages={3975--3985},
  year={2025}
}

@article{liu2025surveillancevqa,
  title={SurveillanceVQA-589K: A Benchmark for Comprehensive Surveillance Video-Language Understanding with Large Models},
  author={Liu, Bo and Qiao, Pengfei and Ma, Minhan and Zhang, Xuange and Tang, Yinan and Xu, Peng and Liu, Kun and Yuan, Tongtong},
  journal={arXiv preprint arXiv:2505.12589},
  year={2025}
}

@article{kim2025vru,
  title={VRU-Accident: A Vision-Language Benchmark for Video Question Answering and Dense Captioning for Accident Scene Understanding},
  author={Kim, Younggun and Abdelrahman, Ahmed S and Abdel-Aty, Mohamed},
  journal={arXiv preprint arXiv:2507.09815},
  year={2025}
}

@article{gao2025vagu,
  title={VAGU \& GtS: LLM-Based Benchmark and Framework for Joint Video Anomaly Grounding and Understanding},
  author={Gao, Shibo and Yang, Peipei and Liu, Yangyang and Chen, Yi and Zhu, Han and Zhang, Xuyao and Huang, Linlin},
  journal={arXiv preprint arXiv:2507.21507},
  year={2025}
}

@article{zhang2025qwen3-embedding,
  title={Qwen3 Embedding: Advancing Text Embedding and Reranking Through Foundation Models},
  author={Zhang, Yanzhao and Li, Mingxin and Long, Dingkun and Zhang, Xin and Lin, Huan and Yang, Baosong and Xie, Pengjun and Yang, An and Liu, Dayiheng and Lin, Junyang and others},
  journal={arXiv preprint arXiv:2506.05176},
  year={2025}
}

@inproceedings{reimers-2019-sentence-bert,
    title = "Sentence-BERT: Sentence Embeddings using Siamese BERT-Networks",
    author = "Reimers, Nils and Gurevych, Iryna",
    booktitle = "Proceedings of the 2019 Conference on Empirical Methods in Natural Language Processing",
    month = "11",
    year = "2019",
    publisher = "Association for Computational Linguistics",
    url = "https://arxiv.org/abs/1908.10084",
}

@inproceedings{huangexvad,
  title={Ex-VAD: Explainable Fine-grained Video Anomaly Detection Based on Visual-Language Models},
  author={Huang, Chao and Shi, Yushu and Wen, Jie and Wang, Wei and Xu, Yong and Cao, Xiaochun},
  booktitle={Forty-second International Conference on Machine Learning},
  year={2025}
}

@inproceedings{cai2025hiprobe,
  title={Hiprobe-vad: Video anomaly detection via hidden states probing in tuning-free multimodal llms},
  author={Cai, Zhaolin and Li, Fan and Zheng, Ziwei and Qin, Yanjun},
  booktitle={Proceedings of the 33rd ACM International Conference on Multimedia},
  pages={592--601},
  year={2025}
}

@article{huang2025vadr1,
  title={Vad-R1: Towards Video Anomaly Reasoning via Perception-to-Cognition Chain-of-Thought},
  author={Huang, Chao and Wang, Benfeng and Wen, Jie and Liu, Chengliang and Wang, Wei and Shen, Li and Cao, Xiaochun},
  journal={Advances in neural information processing systems},
  year={2025}
}

@article{li2025vadtree,
  title={VADTree: Explainable Training-Free Video Anomaly Detection via Hierarchical Granularity-Aware Tree},
  author={Li, Wenlong and Xu, Yifei and Rao, Yuan and Wang, Zhenhua and Deng, Shuiguang},
  journal={arXiv preprint arXiv:2510.22693},
  year={2025}
}

@inproceedings{shao2025eventvad,
  title={Eventvad: Training-free event-aware video anomaly detection},
  author={Shao, Yihua and He, Haojin and Li, Sijie and Chen, Siyu and Long, Xinwei and Zeng, Fanhu and Fan, Yuxuan and Zhang, Muyang and Yan, Ziyang and Ma, Ao and others},
  booktitle={Proceedings of the 33rd ACM International Conference on Multimedia},
  pages={2586--2595},
  year={2025}
}

@inproceedings{chen2025aligning,
  title={Aligning Effective Tokens with Video Anomaly in Large Language Models},
  author={Chen, Yingxian and Liu, Jiahui and Fan, Ruidi and Li, Yanwei and Chang, Chirui and Zhao, Shizhen and Fok, Wilton WT and Qi, Xiaojuan and Wu, Yik-Chung},
  booktitle={Proceedings of the IEEE/CVF International Conference on Computer Vision},
  pages={22695--22706},
  year={2025}
}

@article{zhu2025vau,
  title={VAU-R1: Advancing Video Anomaly Understanding via Reinforcement Fine-Tuning},
  author={Zhu, Liyun and Chen, Qixiang and Shen, Xi and Cun, Xiaodong},
  journal={arXiv preprint arXiv:2505.23504},
  year={2025}
}

@article{yu2025cuebench,
  title={CueBench: Advancing Unified Understanding of Context-Aware Video Anomalies in Real-World},
  author={Yu, Yating and Cao, Congqi and Wang, Zhaoying and Meng, Weihua and Li, Jie and Li, Yuxin and Wei, Zihao and Shen, Zhongpei and Zhang, Jiajun},
  journal={arXiv preprint arXiv:2511.00613},
  year={2025}
}

@article{mo2025a2seek,
  title={A2Seek: Towards Reasoning-Centric Benchmark for Aerial Anomaly Understanding},
  author={Mo, Mengjingcheng and Tong, Xinyang and Tan, Mingpi and Leng, Jiaxu and Zheng, Jiankang and Liu, Yiran and Chen, Haosheng and Gan, Ji and Li, Weisheng and Gao, Xinbo},
  journal={arXiv preprint arXiv:2505.21962},
  year={2025}
}

@article{zhan2025l2v,
  title={L2V-CoT: Cross-Modal Transfer of Chain-of-Thought Reasoning via Latent Intervention},
  author={Zhan, Yuliang and Tang, Xinyu and Wan, Han and Li, Jian and Wen, Ji-Rong and Sun, Hao},
  journal={arXiv preprint arXiv:2511.17910},
  year={2025}
}

@article{liu2025survqa,
  title={SurveillanceVQA-589K: A Benchmark for Comprehensive Surveillance Video-Language Understanding with Large Models},
  author={Liu, Bo and Qiao, Pengfei and Ma, Minhan and Zhang, Xuange and Tang, Yinan and Xu, Peng and Liu, Kun and Yuan, Tongtong},
  journal={arXiv preprint arXiv:2505.12589},
  year={2025}
}

@article{zhang2019bertscore,
  title={Bertscore: Evaluating text generation with bert},
  author={Zhang, Tianyi and Kishore, Varsha and Wu, Felix and Weinberger, Kilian Q and Artzi, Yoav},
  journal={arXiv preprint arXiv:1904.09675},
  year={2019}
}

@inproceedings{sellam2020bleurt,
  title={BLEURT: Learning robust metrics for text generation},
  author={Sellam, Thibault and Das, Dipanjan and Parikh, Ankur},
  booktitle={Proceedings of the 58th annual meeting of the association for computational linguistics},
  pages={7881--7892},
  year={2020}
}
}

\begin{IEEEbiography}[{\includegraphics[width=1in,height=1in,clip,keepaspectratio]{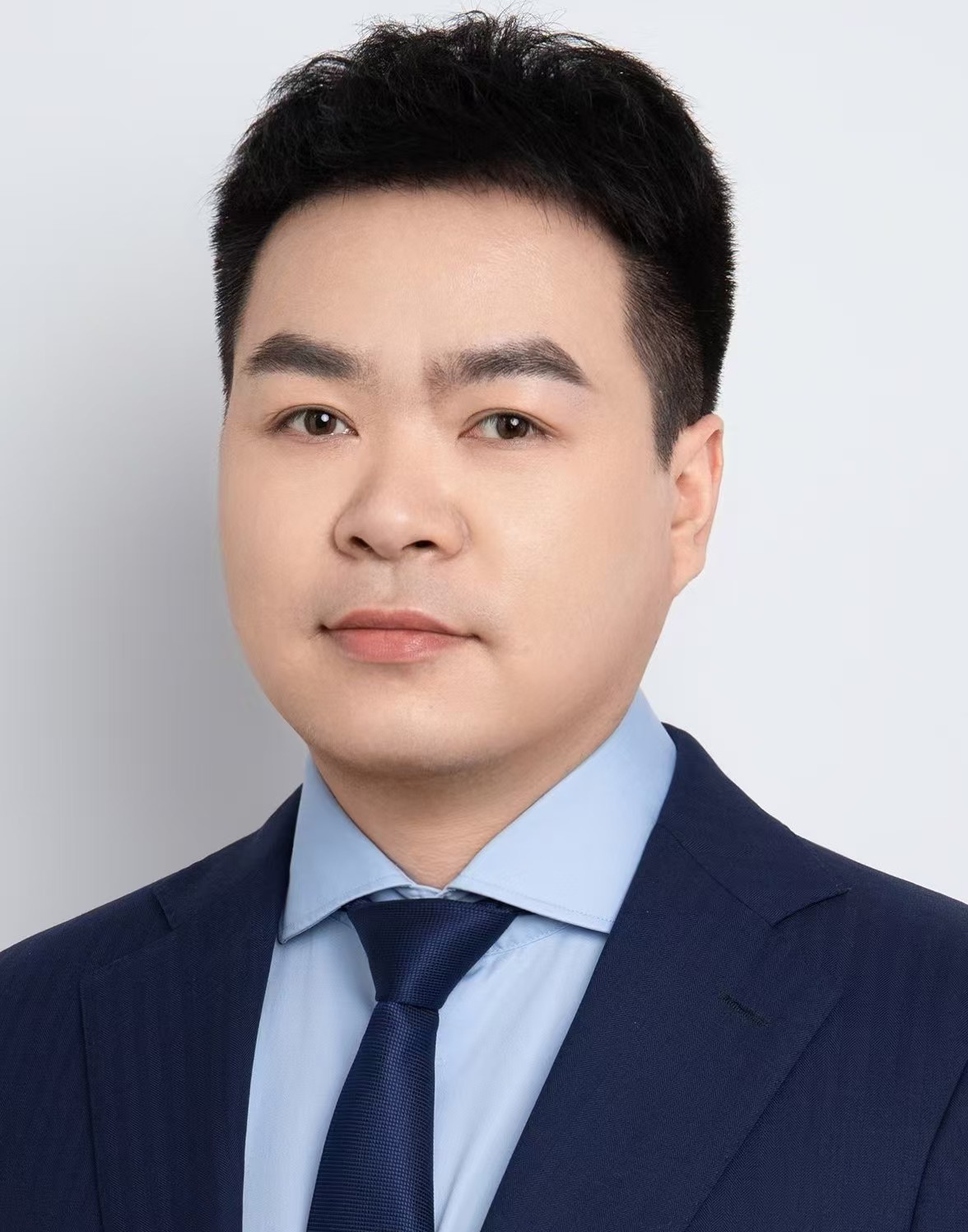}}]{Chao Huang} (Member, IEEE) received the Ph.D. degree in computer science and technology from Harbin Institute of Technology, Shenzhen, China, in 2022. From 2019 to 2022, he was a visiting scholar with Peng Cheng Laboratory, Shenzhen. He is currently an Assistant Professor with the School of Cyber Science and Technology, Sun Yat-sen University, Shenzhen. So far, he has published over 70 top-tier journal and conference papers in prestigious international journals and conferences. His research interests include anomaly detection, multimedia analysis, object detection, image/video compression, and deep learning. Dr. Huang received the Distinguished Paper Award of AAAI 2023, and his dissertation was nominated for Harbin Institute of Technology’s Outstanding Dissertation Award. He serves as an Associate Editor for `\textit{IEEE Transactions on Image Processing}', `\textit{Pattern Recognition}' and serves/served as the reviewer/PC member for several top-tier journals and conferences, including IEEE TPAMI, TIP, TIFS, ACM CSUR, CVPR, ICCV, ECCV, ICML, NeurIPS, ICLR, AAAI, IJCAI, and ACM Multimedia.
\end{IEEEbiography}

\vspace{-1cm}

\begin{IEEEbiography}[{\includegraphics[width=1in,height=1in,clip,keepaspectratio]{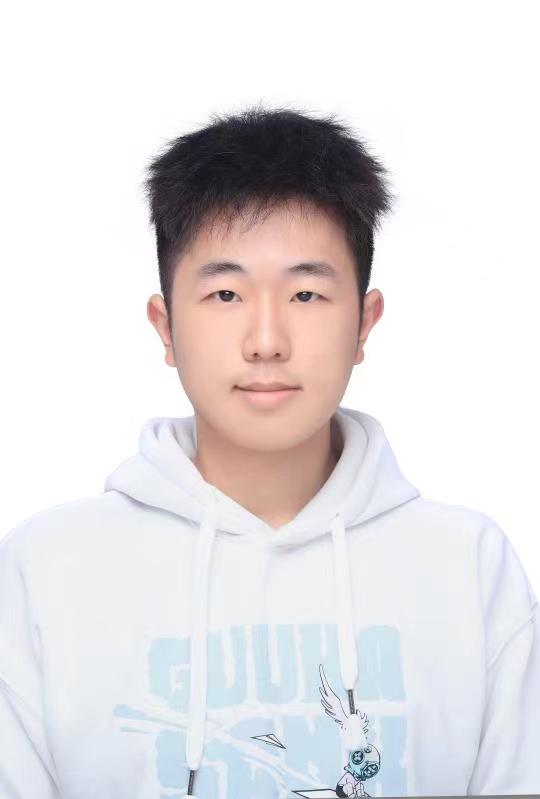}}]{Benfeng Wang} received the B.E. degree from the School of Computer Science and Engineering, Central South University, Changsha, China, in 2024. He is currently pursuing the M.S. degree with the School of Cyber Science and Technology, Sun Yat-sen University, Shenzhen, China. His research interests include multimodal large language models, video analysis, and anomaly detection.
\end{IEEEbiography}

\vspace{-1cm}

\begin{IEEEbiography}[{\includegraphics[width=1in,height=1in,clip,keepaspectratio]{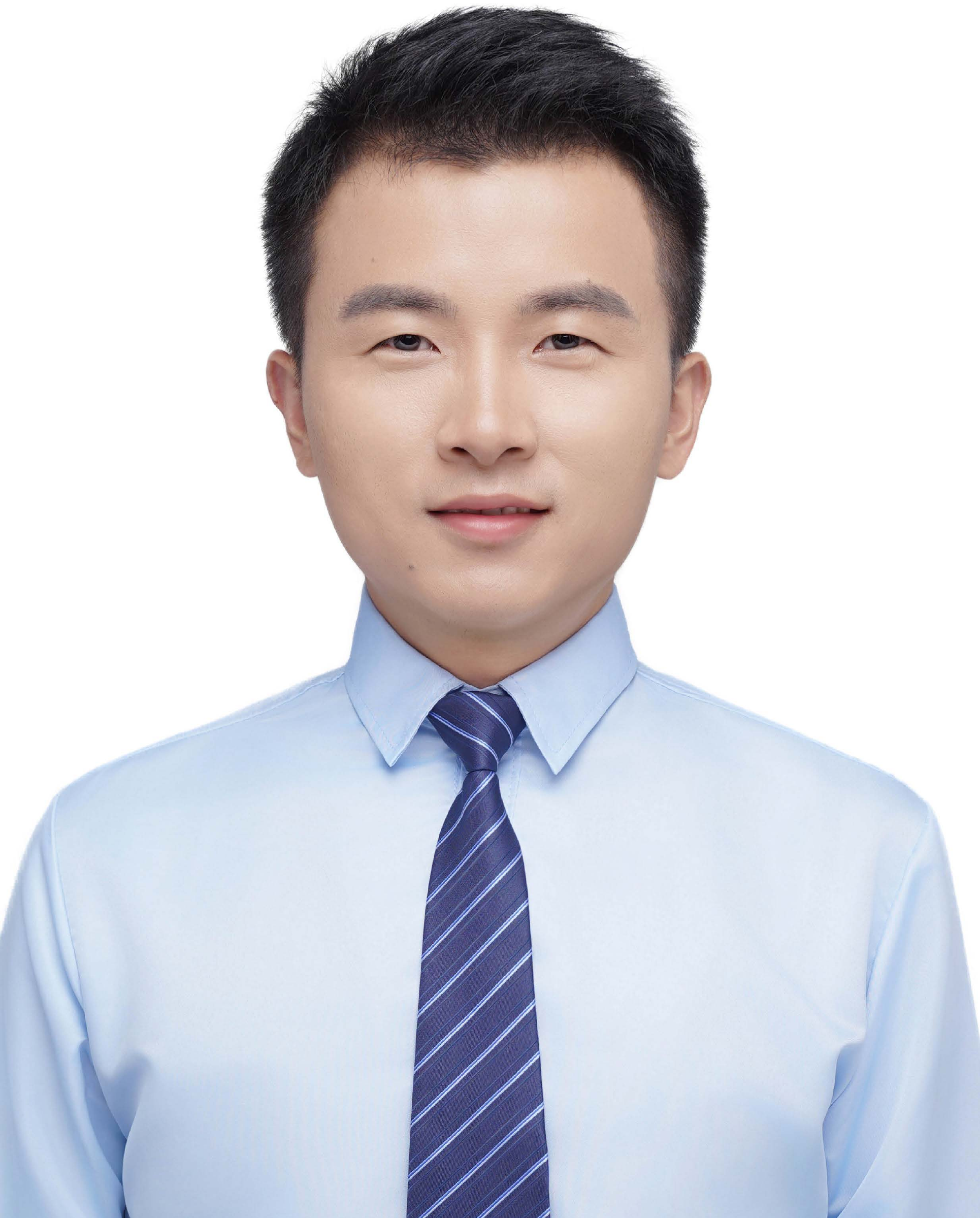}}]{Wei Wang} is currently a post-doctoral candidate at the School of Cyber Science and Technology, Shenzhen Campus of Sun Yat-Sen University, Shenzhen, China. He received the Ph.D. degree at the School of Software Technology, Dalian University of Technology, Dalian, China, in 2022. He received the M.S. degree at the School of Computer Science and Technology from the Anhui University, Hefei, China, in 2018. He received the B.S. degree at the School of Science from the Anhui Agricultural University, Hefei, China, in 2015. His major research interests include transfer learning, zero-shot learning, deep learning, etc.  
\end{IEEEbiography}

\vspace{-1cm}

\begin{IEEEbiography}[{\includegraphics[width=1in,height=1in,clip,keepaspectratio]{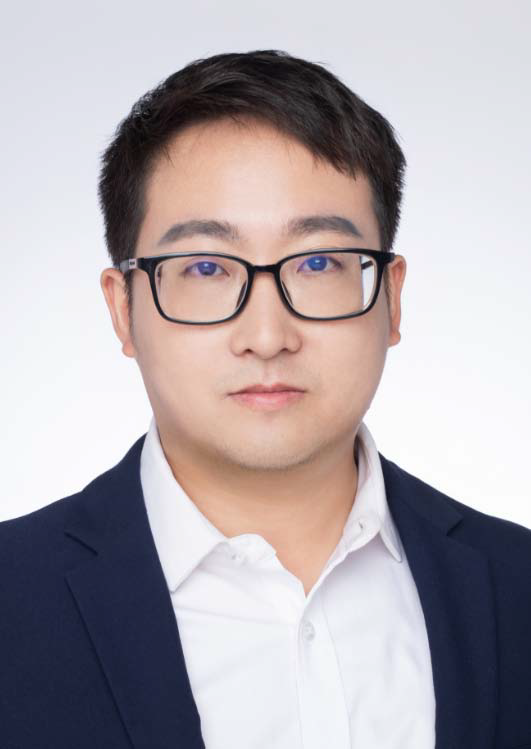}}]{Jie Wen} (Senior Member, IEEE) received the Ph.D. degree in Computer Science and Technology at Harbin Institute of Technology, Shenzhen in 2019. He is currently an Associate Professor at the School of Computer Science and Technology, Harbin Institute of Technology, Shenzhen. His research interests include image and video enhancement, pattern recognition, and machine learning. He has authored or co-authored more than 100 technical papers at prestigious international journals and conferences, including the TNNLS, TIP, TCYB, NeurIPS, ICML, CVPR, AAAI, IJCAI, ACM MM, etc. He serves as an Associate Editor of IEEE Transactions on Pattern Analysis and Machine Intelligence, IEEE Transactions on Image Processing, IEEE Transactions on Information Forensics and Security, Pattern Recognition, and International Journal of Image and Graphics, an Area Editor of Information Fusion. He also served as the Area Chair of ACM MM and ICML. He was selected for the ‘World’s Top 2\% Scientists List’ in 2021-2024. One paper received the ‘distinguished paper award’ from AAAI’23. 
\end{IEEEbiography}

\vspace{-1cm}

\begin{IEEEbiography}
[{\includegraphics[width=1in,height=1in,clip,keepaspectratio]{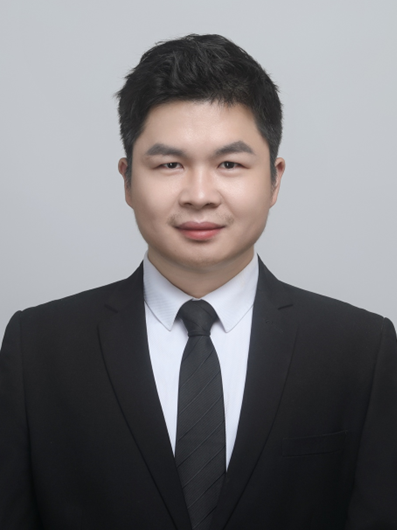}}]
{Li Shen} is currently an associate professor at Sun Yat-sen University. Previously, he was a research scientist at JD Explore Academy, Beijing, and a senior researcher at Tencent AI Lab, Shenzhen. He received his bachelor's degree and Ph.D. from the School of Mathematics, South China University of Technology. His research interests include theory and algorithms for nonsmooth convex and nonconvex optimization, and their applications in trustworthy artificial intelligence, deep learning, and reinforcement learning. He has published more than 100 papers in peer-reviewed top-tier journal papers (JMLR, IEEE TPAMI, IJCV, IEEE TSP, IEEE TIP, IEEE TKDE, etc.) and conference papers (ICML, NeurIPS, ICLR, CVPR, ICCV, etc.). He serves as an Associate Editor of IEEE Transactions on Pattern Analysis and Machine Intelligence. 
\end{IEEEbiography}

\vspace{-1cm}

\begin{IEEEbiography}[{\includegraphics[width=1in,height=1in,clip,keepaspectratio]{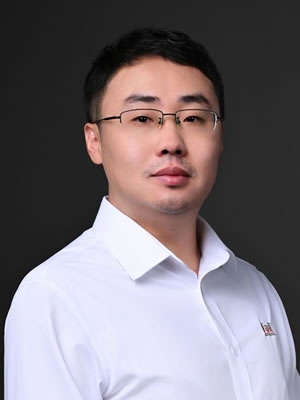}}]{Wenqi Ren} (Member, IEEE) received the PhD degree from Tianjin University, Tianjin, China, in 2017. From 2015 to 2016, he was supported by the China Scholarship Council and working with Prof. MingHusan Yang as a Joint-Training PhD Student with the Electrical Engineering and Computer Science Department, University of California at Merced. He is currently a professor with the School of Cyber Science and Technology, Shenzhen Campus, Sun Yat-sen University, Shenzhen, China. His research interests include image processing and related high-level vision problems. He received the Tencent Rhino Bird Elite Graduate Program Scholarship in 2017 and the MSRA Star Track Program in 2018.

\end{IEEEbiography}

\begin{IEEEbiography}[{\includegraphics[width=1in,height=1in,clip,keepaspectratio]{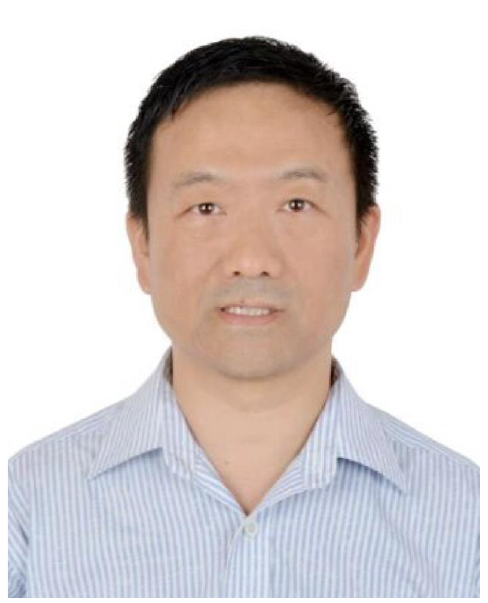}}]{Yong Xu}
(Senior Member, IEEE) received his B.S. degree, M.S. degree in 1994 and 1997, respectively. He received the Ph.D. degree in Pattern Recognition and Intelligence system at NUST (China) in 2005. He is currently a Professor with the School of Computer Science and Technology, Harbin Institute of Technology (HIT), Shenzhen. His research interests include pattern recognition, deep learning, biometrics, machine learning and video analysis. He has published over 70 papers in top-tier academic journals and conferences. His articles have been cited more than 8,000 times in the Web of Science, and 27,000 times in the Google Scholar. He has served as a Co-Editors-in-Chief of the International Journal of Image and Graphics, an Associate Editor of the CAAI Transactions on Intelligence Technology, an editor of the Pattern Recognition and Artificial Intelligence. 
\end{IEEEbiography}

\begin{IEEEbiography}[{\includegraphics[width=1in,height=1.2in,clip,keepaspectratio]{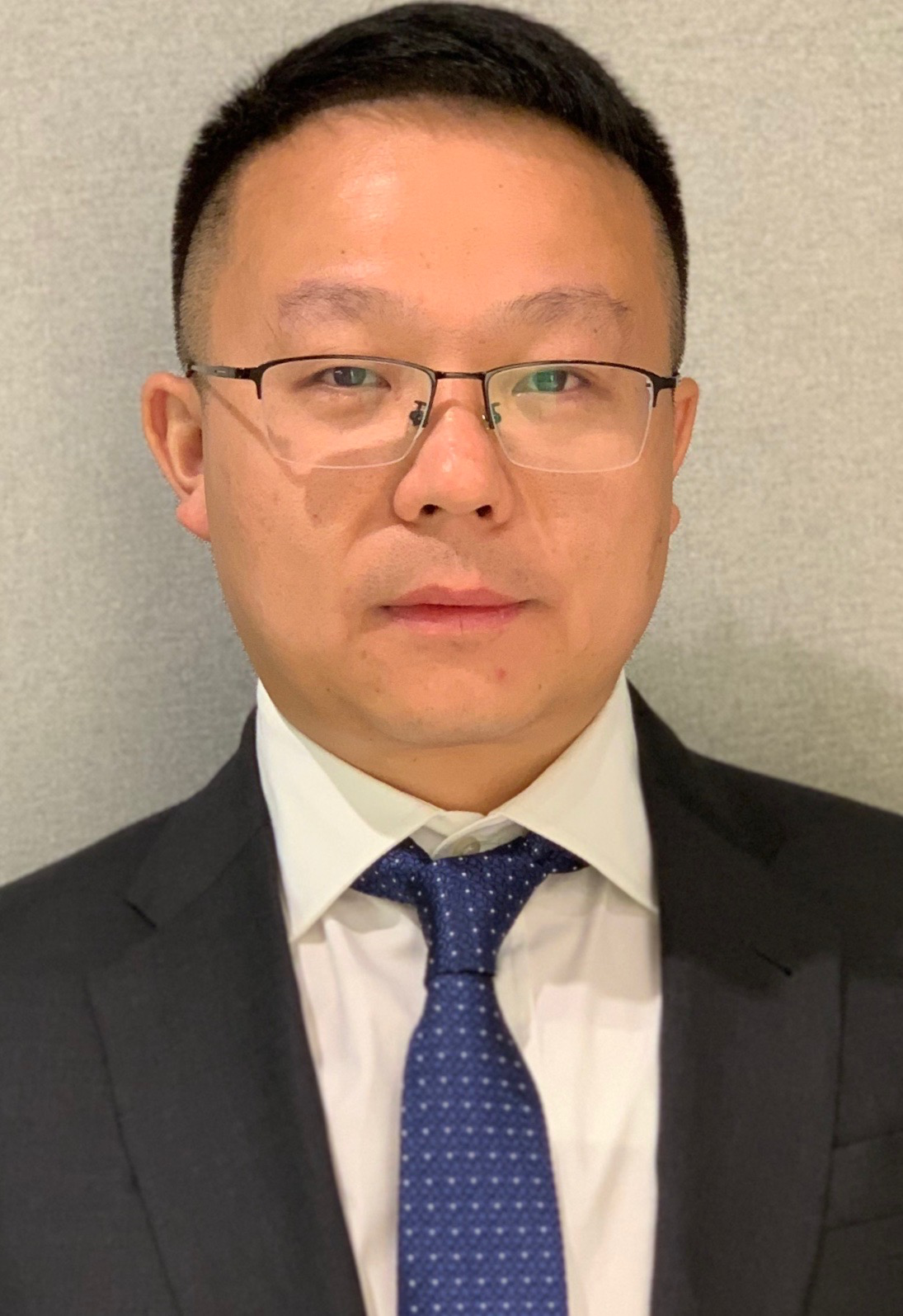}}]{Xiaochun Cao} (Senior Member, IEEE)  Cao is a Professor and Dean of the School of Cyber Science and Technology, Shenzhen Campus of Sun Yat-sen University. He received the B.E. and M.E. degrees both in computer science from Beihang University (BUAA), China, and the Ph.D. degree in computer science from the University of Central Florida, USA, with his dissertation nominated for the university-level Outstanding Dissertation Award. After graduation, he spent about three years at ObjectVideo Inc. as a Research Scientist. From 2008 to 2012, he was a professor at Tianjin University. Before joining SYSU, he was a professor at the Institute of Information Engineering, Chinese Academy of Sciences. He has authored and coauthored over 200 journal and conference papers. In 2004 and 2010, he was the recipient of the Piero Zamperoni best student paper award at the International Conference on Pattern Recognition. He is on the editorial boards of IEEE Transactions on Pattern Analysis and Machine Intelligence and IEEE Transactions on Image Processing, and was on the editorial boards of IEEE Transactions on Circuits and Systems for Video Technology and IEEE Transactions on Multimedia. 
\end{IEEEbiography}

\end{document}